%% file: main.tex
\documentclass[10pt,letterpaper,twocolumn]{article}
\usepackage{mercor-research}
\ResearchSetHeadlineFont{inter}

\usepackage{paper-style}

\usepackage{pgfplots}
\pgfplotsset{compat=1.18}
\usetikzlibrary{arrows.meta,decorations.pathreplacing,calc}
\usepackage{float}
\usepackage[backend=biber,style=numeric,sorting=none,natbib=true]{biblatex}
\usepackage[capitalize,noabbrev]{cleveref}
\newcommand{\bench}{OWE-Bench}
\newcommand{\numtasks}{26}

\newif\ifdraft \draftfalse

\newcommand{\Tnowval}{1~October~2026}
\input{figures/search_numbers}

\input{figures/results_numbers}

\title{\bench: A long-horizon benchmark for measuring real-world engineering}
\author{Jacob Cole \and Martin Dimakis \and Victor Barres \and Cindy Yang \and Aravind Selvan \and Daniel Babalola \and Bertie Vidgen \and Edward J. Hu \and Jason Wei}
\ResearchSetAffiliation{Mercor}
\ResearchSetShortTitle{\bench}
\ResearchSetDate{November 2026}
\ResearchSetContactEmail{research@mercor.com}
\ResearchSetCodeURL{https://github.com/Mercor-Intelligence/OpenWorldEngineer}
\ResearchSetDataURL{https://huggingface.co/datasets/mercor/OpenWorldEngineer}
\ResearchSetLeaderboardURL{https://openworldengineer.mercor.com}
\hypersetup{pdftitle={OWE-Bench: A long-horizon benchmark for measuring real-world engineering}}

\begin{document}
\twocolumn[{%
\ResearchMakeTitle

\begin{researchabstract}
We present \bench ~(OpenWorldEngineer-Bench), a benchmark of 26 long-horizon engineering tasks written by domain experts in software engineering, physical engineering, machine learning, and scientific computing. Tasks range from optimizing kernels and designing an aircraft, to identifying the faults responsible for an earthquake. Each task requires significant domain knowledge and engineering work, with models averaging hundreds of millions of tokens per task. Solutions are scored on practical measures such as speed, cost and accuracy, both on provided cases the agent could test against and on hidden cases that an engineer would expect from the task description. The best-performing frontier model so far, Claude Fable 5.2, scores 54.6\% and averages 211 million tokens per task. We highlight two findings. First, when models are given more time, they use more test-time compute and achieve higher scores. Models also scale differently with test-time compute: GPT-6 Astra achieves higher scores at shorter time limits and at equal token use, while Claude Fable 5.2 uses more tokens at longer limits to achieve the highest overall score. Second, each model scores lower on hidden cases than on the provided cases, by an average of 18 points, which we call the deployment gap.  Models also fail to predict the deployment gap, overestimating their performance on hidden test cases by an average of 21 points. We hope that \bench{} grounds progress toward agents whose engineering solutions perform as well in deployment as they do in testing.
\end{researchabstract}

\begin{minipage}{\textwidth}\centering\captionsetup{hypcap=false}\input{figures/teaser_time_scaling}\end{minipage}\par
\vspace{14pt}
}]

\input{sections/introduction}
\input{sections/principles}
\input{sections/benchmark}

\input{sections/construction}
\input{sections/results}
\input{sections/analysis}
\input{sections/cheating}
\input{sections/discussion}

\printbibliography[heading=bibintoc]

\ResearchAppendixOnNewPage
\appendix
\twocolumn[{\ResearchAppendixContents}]
\input{sections/appendix_tasks}
\input{sections/appendix_dev}
\input{sections/appendix_search}
\input{sections/appendix_exploits}

\input{sections/appendix_domain_results}
\input{sections/appendix_leaderboard}
\input{sections/related_work}

\end{document}

%% file: figures/search_numbers.tex
\newcommand{\SearchDelta}{$+0.003$}
\newcommand{\SearchCI}{$[-0.011,0.018]$}
\newcommand{\SearchUptake}{13\%}

\newcommand{\MatchedPooledDelta}{$0.000$}

\newcommand{\MatchedPooledCI}{$[-0.035,0.035]$}

%% file: figures/results_numbers.tex
\newcommand{\LBdate}{11~November~2026}

\newcommand{\LBscheduled}{2,912}

\newcommand{\TopGap}{45.4}
\newcommand{\NextThreeSpread}{3.0}
\newcommand{\BestGainMin}{13.7}
\newcommand{\BestGainMax}{18.3}
\newcommand{\MeanFableTwo}{54.6}
\newcommand{\LowFableTwo}{46.2}
\newcommand{\HighFableTwo}{63.0}
\newcommand{\BestFableTwo}{68.3}
\newcommand{\CostFableTwo}{105.08}
\newcommand{\TokensFableTwo}{210.7}

\newcommand{\MeanAstra}{44.2}

\newcommand{\BestAstra}{58.3}
\newcommand{\CostAstra}{234.63}
\newcommand{\TokensAstra}{88.8}

\newcommand{\MeanOpus}{41.6}

\newcommand{\BestOpus}{58.4}

\newcommand{\MeanSol}{41.2}

\newcommand{\CostSol}{52.12}

\newcommand{\TimeSol}{95}

\newcommand{\MeanSonnet}{39.0}

\newcommand{\CostSonnet}{38.65}

\newcommand{\MeanCarbon}{32.8}

\newcommand{\MeanFable}{29.1}

\newcommand{\CostFable}{87.70}

\newcommand{\TimeFable}{65}

\newcommand{\TimeInkling}{8}

\newcommand{\DomSecondSWE}{49.7}

\newcommand{\DomSecondPE}{57.3}

\newcommand{\DomSecondML}{53.2}

\newcommand{\DomSecondSC}{35.7}
\newcommand{\DomMarginMin}{1.8}
\newcommand{\DomMarginMax}{12.5}
\newcommand{\LeadGap}{10.4}
\newcommand{\LeadGapLow}{3.7}
\newcommand{\LeadGapHigh}{17.1}
\newcommand{\LeadTasks}{16}

%% file: figures/teaser_time_scaling.tex
\centering
\begin{minipage}{\textwidth}
{\centering\includegraphics[width=0.874\linewidth]{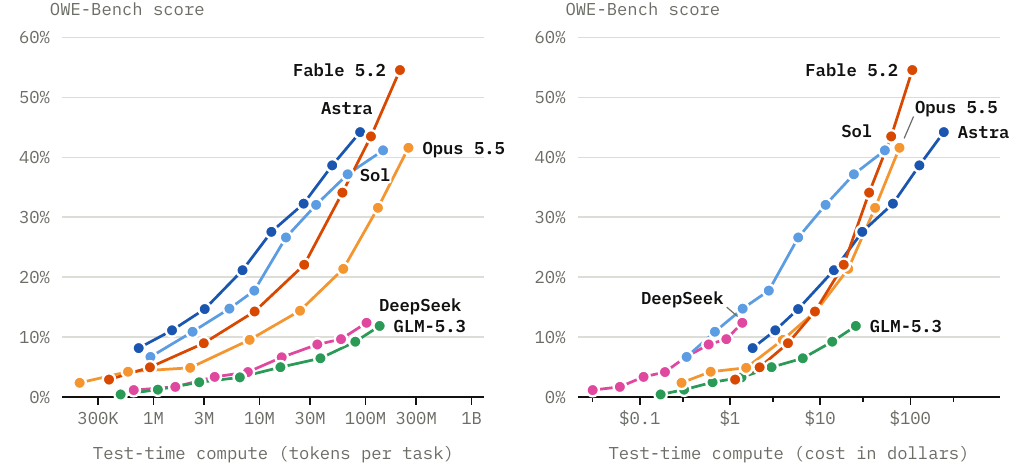}\par}
\captionof{figure}{\textbf{Every model scores higher with more time.} OWE-Bench score against tokens (left) and model cost (right) per task for six models. Each line's dots go from 1/128 of every task's time limit to the full limit, doubling at each step (four attempts per task, and the leaderboard's eight at the full limit). Points average all 26 tasks. Tokens count input plus output and costs are model-usage charges at list prices, as in \cref{fig:efficiency}; horizontal axes use log scales.}
\label{fig:time-scaling}
\end{minipage}

%% file: sections/introduction.tex
\section{Introduction}
\label{sec:intro}

Frontier AI agents can now carry out substantial engineering work end to end \citep{kwa2025timehorizon,wijk2024rebench,rank2026posttrainbench,frontierswe}. Given a bug report, an optimization target or a design goal, an agent can plan an approach, implement it, test it and revise its solution without step-by-step human direction. Prior to deployment, a human is still expected to decide whether the agent's work is ready to use. For agents to assume greater responsibility, they must be able to make this judgment themselves. Much of the difficulty in real engineering stems from the fact that a goal can be clearly stated while also leaving certain decisions up for judgment. Those decisions include which conditions a solution must handle and how much evidence is enough to show it is ready. An engineer must decide what to test, whether observed improvements will hold beyond those tests, which version to deliver, and whether further work is worth its cost. We ask: \emph{can agents tell when the job is done?}

Evaluating this ability requires distinguishing what an agent can establish during development from how its solution ultimately performs. A final score measures the latter, but does not by itself reveal how reliable the evidence available to the agent was. Existing benchmarks evaluate agents on extended workflows, including repository changes, research engineering, model training and paper replication \citep{jimenez2024swebench,merrill2026terminalbench,desai2026swemarathonagentsautonomouslycomplete,chan2024mlebench,starace2025paperbench}, but primarily report the quality of the final artifact. In SWE-bench and Terminal-Bench, hidden pass/fail tests determine success for the evaluator, but do not measure whether the agent could anticipate that outcome before submission. Other engineering benchmarks allow agents to query the grader's own simulation while they work \citep{son2026feacad,zhang2026funcad}, enabling them to optimize directly against it. 

Separating development from evaluation is a basic principle of machine learning \citep{hastie2009elements}, yet agent benchmarks do not always enforce it \citep{kapoor2024agentsmatter}, and when an agent can repeatedly query its final grader, generalization is hard to tell apart from optimization against the benchmark itself \citep{dwork2015reusable,blum2015ladder}. Where final cases are hidden, as in MLE-bench and ALE-Bench \citep{chan2024mlebench,imajuku2025alebench}, benchmarks report how well the final work scores, but not how much of the improvement an agent measured during development survives on conditions it did not see.
\begin{figure}[!t]
  \centering
  \input{figures/teaser_artifacts}
\end{figure}%
\begin{figure*}[t]
  \input{figures/teaser_dev_leaderboard}
\end{figure*}%
We built \bench{} to study engineering agents when the evidence available during development does not fully reveal how their work will perform. It contains \numtasks{} expert-authored tasks across software engineering, physical engineering, machine learning and scientific computing, with 4 to 18 hours per task. Agents optimize database engines and GPU kernels, post-train language models, reconstruct CAD from drawings, and design structures and aircraft. Each task has a continuous performance objective without a predefined threshold for completion. Agents can develop, test and revise solutions using the evidence and tools available in their environments, but their final work is evaluated independently on hidden conditions they cannot inspect or query. Where the provided cases can be scored, we apply the same scoring rule to provided and hidden cases to measure the deployment gap: the difference between performance demonstrated during development and performance on unseen conditions. We call this setting open-world because agents must decide what to test, which solution to deliver, and when to stop without observing all the conditions that determine final performance. Models make the last of these decisions very differently (\cref{fig:time-scaling}). Run at limits from 1/128 of each task's time to the full limit, the GPT models make better use of less time, scoring highest when time is short. The Claude models stop on their own, at about the same share of whatever limit they are given, yet gain more as the limit grows, and Claude Fable 5.2 scores highest at the full limit.

Our contributions are:
\begin{enumerate}
    \item \textbf{A long-horizon benchmark for engineering autonomy.} ~\bench{} comprises \numtasks{} expert-authored tasks, requiring hundreds of millions of tokens to solve, with continuous objectives and independent evaluation on hidden conditions, designed to leave consequential engineering decisions to the agent (\cref{sec:principles,sec:benchmark,sec:construction}). We establish a leaderboard across fourteen frontier models (\cref{sec:results}).
    \item \textbf{A measurement of the deployment gap.} ~By comparing performance on development and hidden cases, and collecting agents' own forecasts, we show that development evidence helps identify stronger solutions but substantially overstates their performance on unseen conditions (\cref{sec:analysis}).
    \item \textbf{A study of how agents use their time.} ~Running six models at varying time limits, we show that every model scores higher with more time, but that models differ in how much of a limit they use and how much more time helps them, so the order of the models changes with the limit. Surprisingly, even when an agent never runs out of time, the amount of time it is given can change how well it performs (\cref{sec:split}).
\end{enumerate}
We release the benchmark and agent trajectories alongside the leaderboard. We review related work in \cref{sec:related}.

%% file: figures/teaser_artifacts.tex
\centering
\begingroup
\definecolor{pumpslate}{HTML}{475569}
\definecolor{pumprule}{HTML}{CBD5E1}
\begin{tikzpicture}[x=1bp,y=-1bp]
\path[use as bounding box] (0,-1) rectangle (234,231.826);
\node[anchor=base, inner sep=0pt, text=mercorink, font=\fontsize{7}{8.5}\selectfont\bfseries, align=center] at (28.125,15.000) {Ground truth};
\node[anchor=base, inner sep=0pt, text=mercorink, font=\fontsize{7}{8.5}\selectfont\bfseries, align=center] at (87.375,15.000) {Claude\\Fable 5.2};
\node[anchor=base, inner sep=0pt, text=mercorink, font=\fontsize{7}{8.5}\selectfont\bfseries, align=center] at (146.625,15.000) {DeepSeek\\V4.1 Flash};
\node[anchor=base, inner sep=0pt, text=mercorink, font=\fontsize{7}{8.5}\selectfont\bfseries, align=center] at (205.875,15.000) {Beam 1.0};
\node[anchor=west, inner sep=0pt, text=mercorink, font=\fontsize{7}{8.5}\selectfont] at (0.000,25.000) {\textbf{(a) Volute}\enspace\textcolor{pumpslate}{}};
\node[anchor=north west,inner sep=0pt] at (0.000,31.000) {\includegraphics[width=56.250bp]{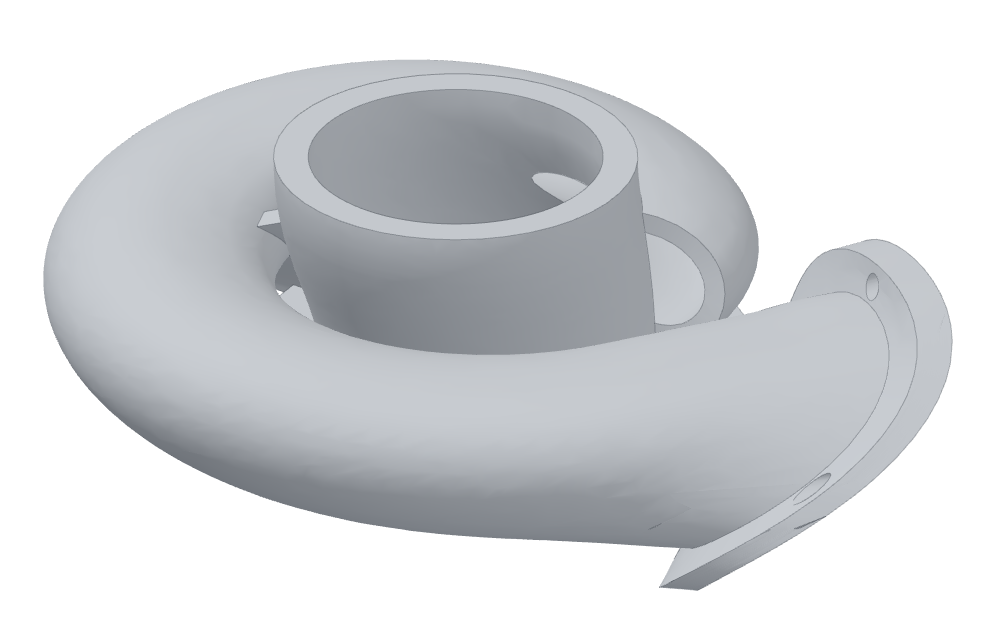}};
\node[anchor=north west,inner sep=0pt] at (59.250,31.000) {\includegraphics[width=56.250bp]{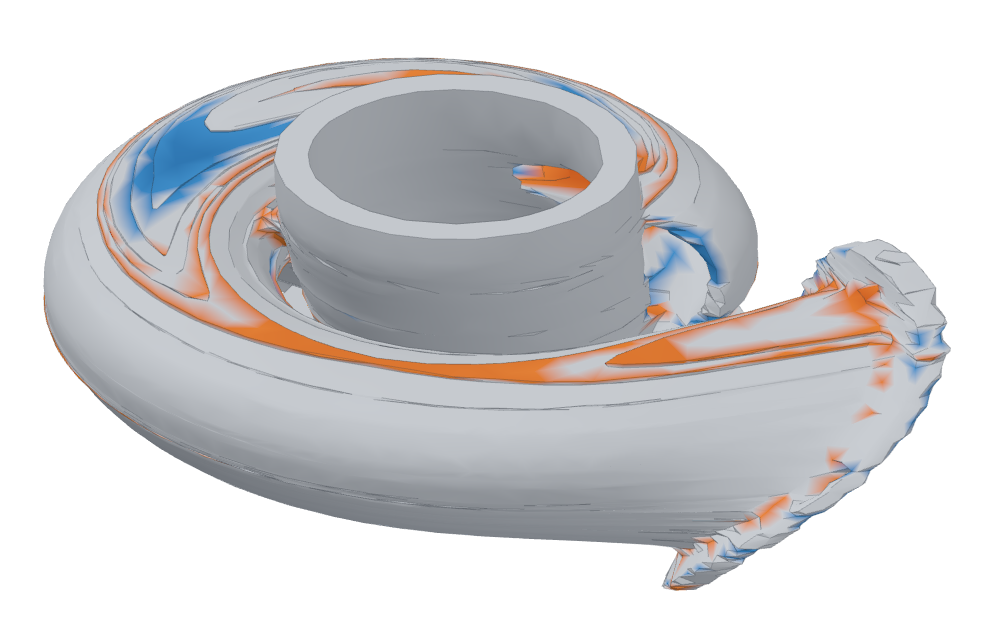}};
\node[anchor=north west,inner sep=0pt] at (118.500,31.000) {\includegraphics[width=56.250bp]{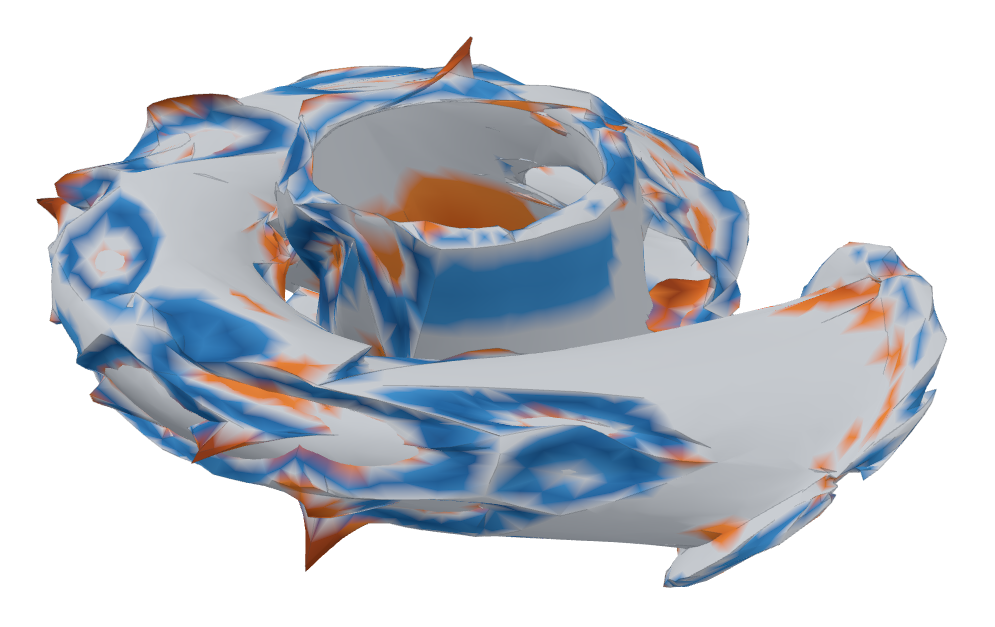}};
\node[anchor=north west,inner sep=0pt] at (177.750,31.000) {\includegraphics[width=56.250bp]{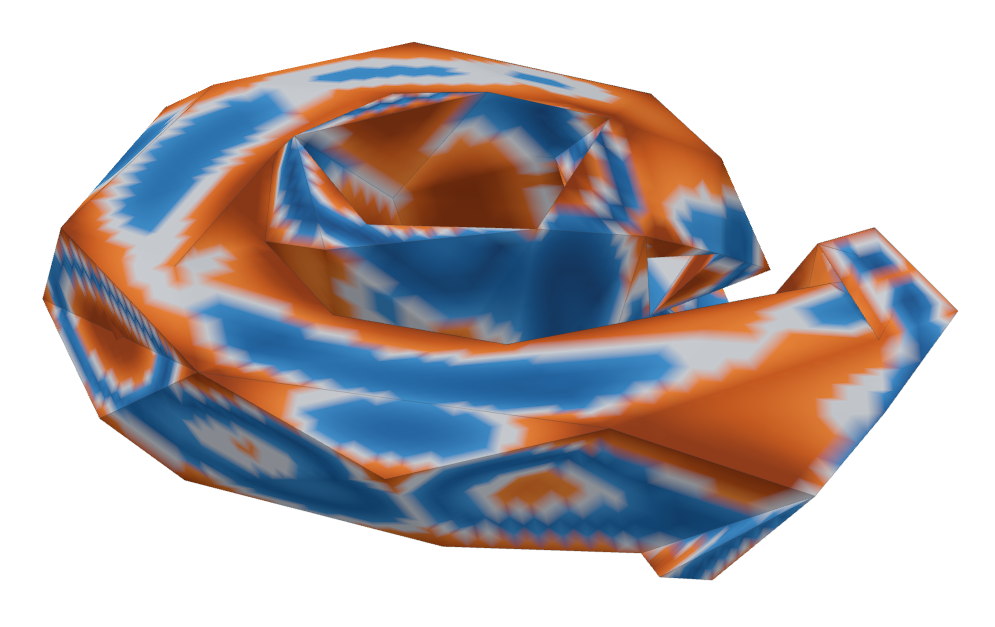}};
\node[anchor=center, inner sep=0pt, text=mercorink, font=\fontsize{7}{8.5}\selectfont] at (87.375,70.446) {85.6\%};
\node[anchor=center, inner sep=0pt, text=mercorink, font=\fontsize{7}{8.5}\selectfont] at (146.625,70.446) {54.0\%};
\node[anchor=center, inner sep=0pt, text=mercorink, font=\fontsize{7}{8.5}\selectfont] at (205.875,70.446) {9.8\%};
\draw[pumprule,line width=0.4bp] (0,78.446) -- (234,78.446);
\node[anchor=west, inner sep=0pt, text=mercorink, font=\fontsize{7}{8.5}\selectfont] at (0.000,85.446) {\textbf{(b) Guide casing}\enspace\textcolor{pumpslate}{}};
\node[anchor=north west,inner sep=0pt] at (0.000,91.446) {\includegraphics[width=56.250bp]{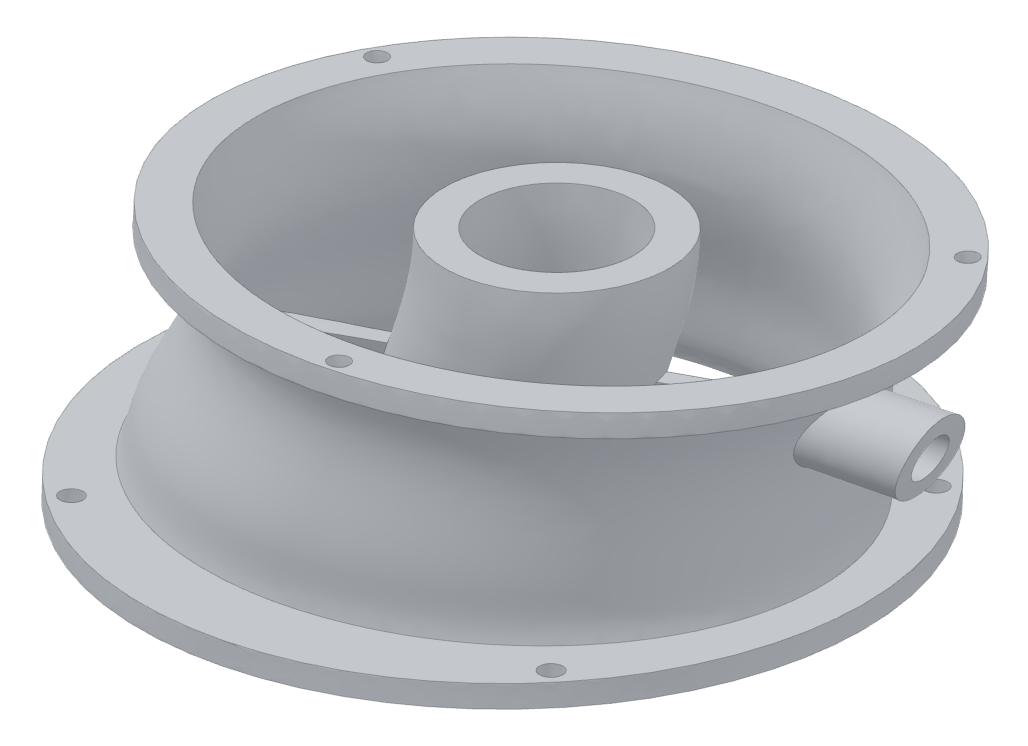}};
\node[anchor=north west,inner sep=0pt] at (59.250,91.446) {\includegraphics[width=56.250bp]{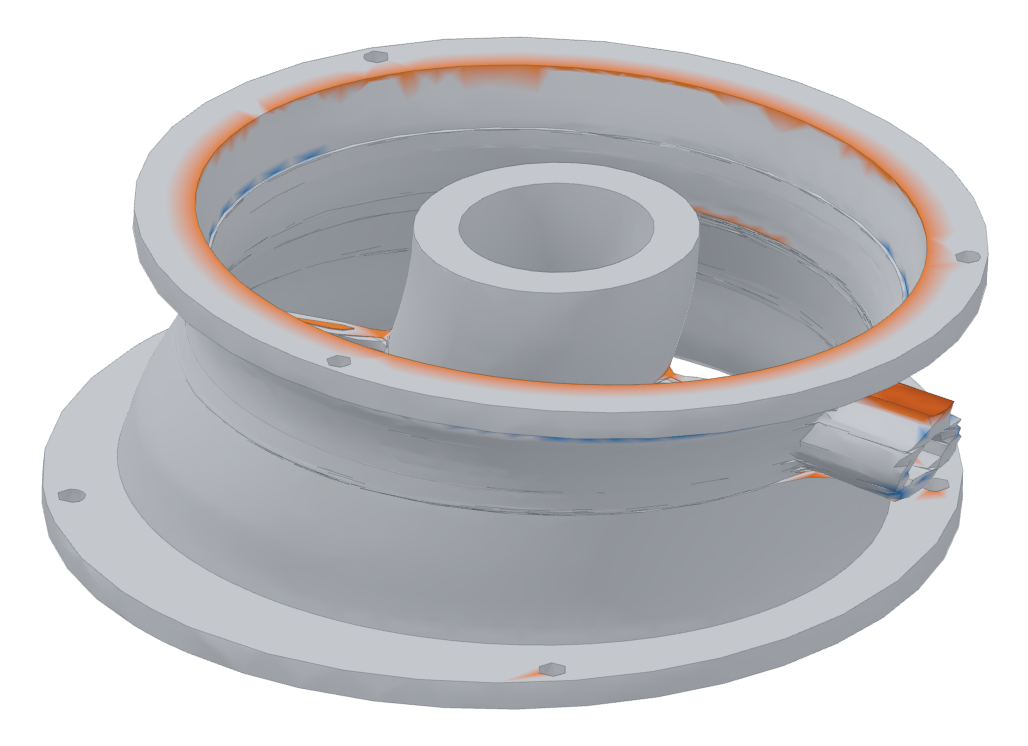}};
\node[anchor=north west,inner sep=0pt] at (118.500,91.446) {\includegraphics[width=56.250bp]{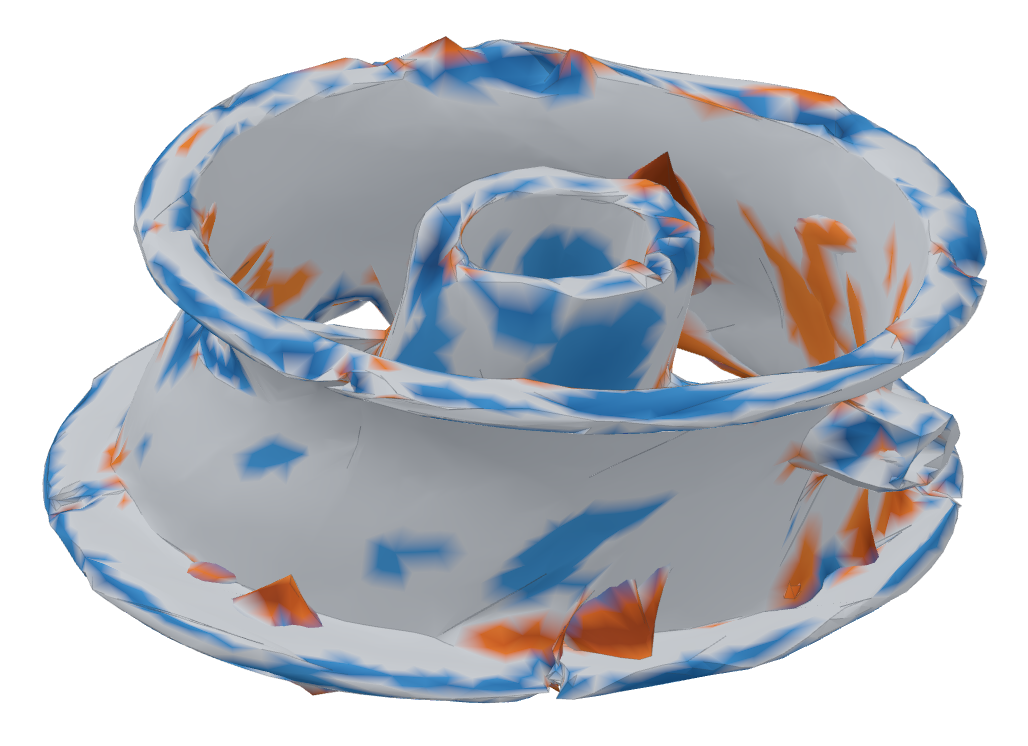}};
\node[anchor=north west,inner sep=0pt] at (177.750,91.446) {\includegraphics[width=56.250bp]{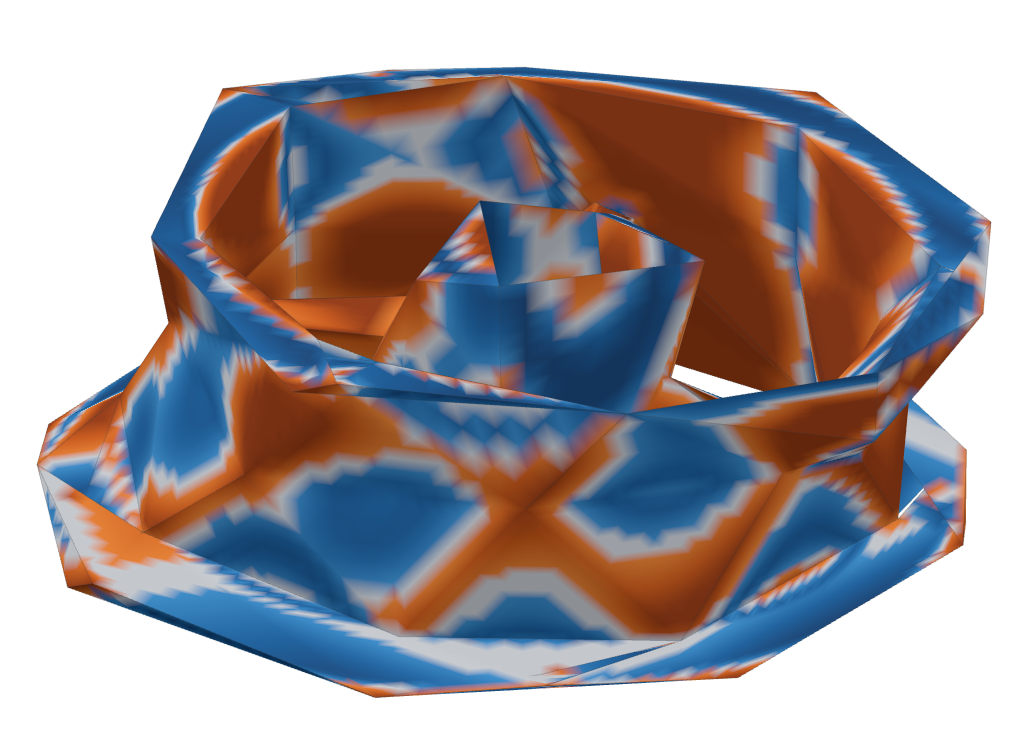}};
\node[anchor=center, inner sep=0pt, text=mercorink, font=\fontsize{7}{8.5}\selectfont] at (87.375,136.266) {95.8\%};
\node[anchor=center, inner sep=0pt, text=mercorink, font=\fontsize{7}{8.5}\selectfont] at (146.625,136.266) {64.3\%};
\node[anchor=center, inner sep=0pt, text=mercorink, font=\fontsize{7}{8.5}\selectfont] at (205.875,136.266) {8.0\%};
\draw[pumprule,line width=0.4bp] (0,144.266) -- (234,144.266);
\node[anchor=west, inner sep=0pt, text=mercorink, font=\fontsize{7}{8.5}\selectfont] at (0.000,151.266) {\textbf{(c) Pump housing}\enspace\textcolor{pumpslate}{}};
\node[anchor=north west,inner sep=0pt] at (0.000,157.266) {\includegraphics[width=56.250bp]{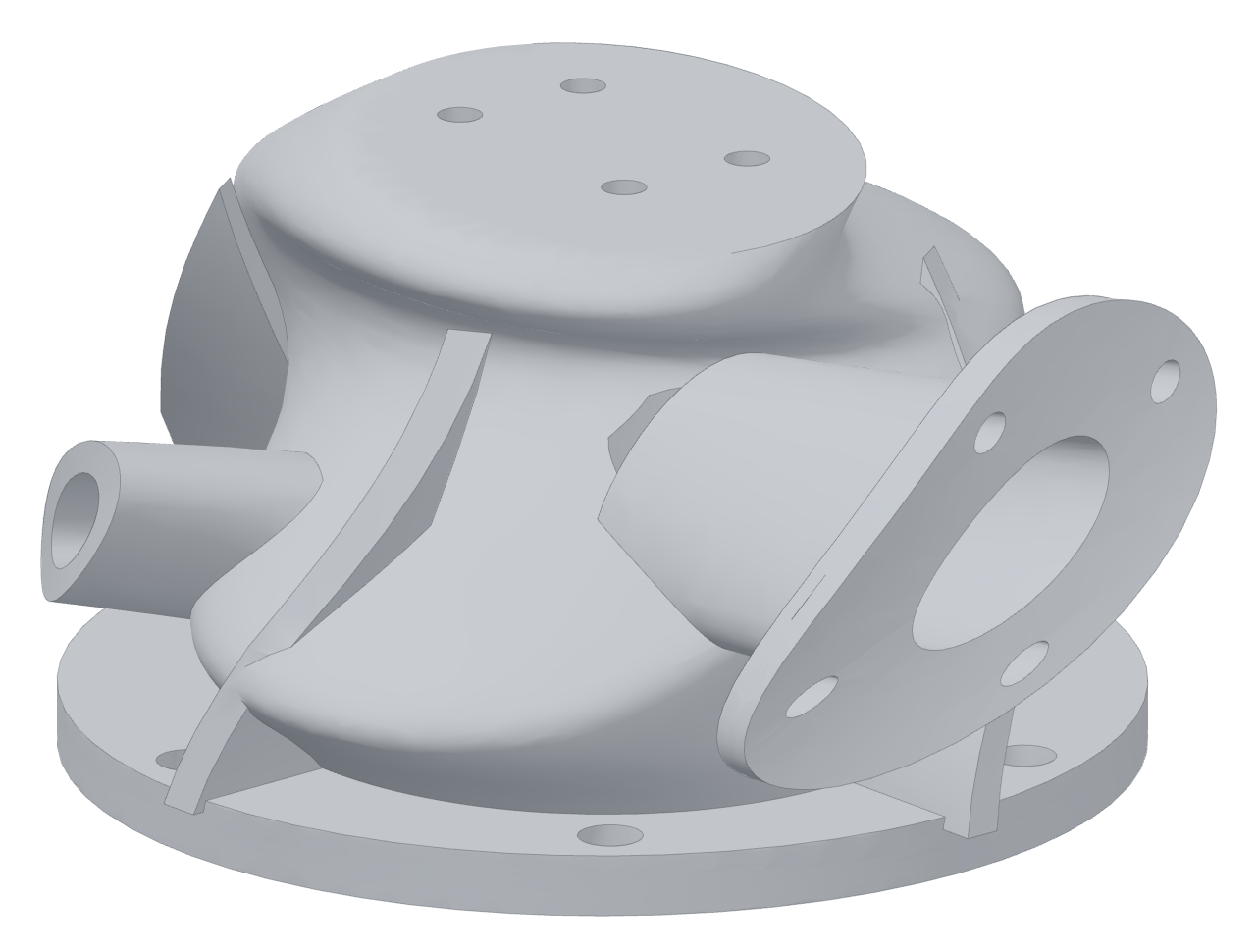}};
\node[anchor=north west,inner sep=0pt] at (59.250,157.266) {\includegraphics[width=56.250bp]{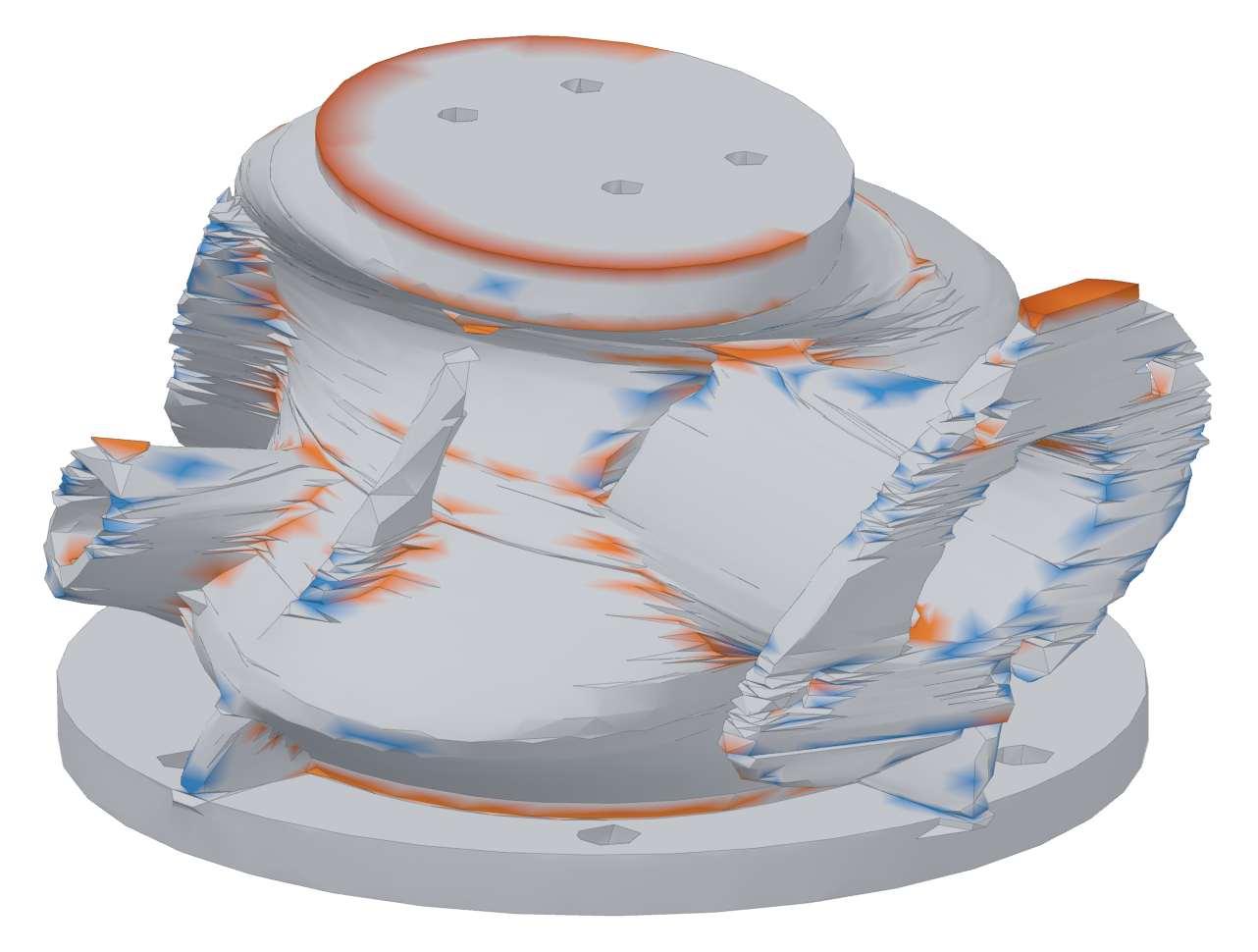}};
\node[anchor=north west,inner sep=0pt] at (118.500,157.266) {\includegraphics[width=56.250bp]{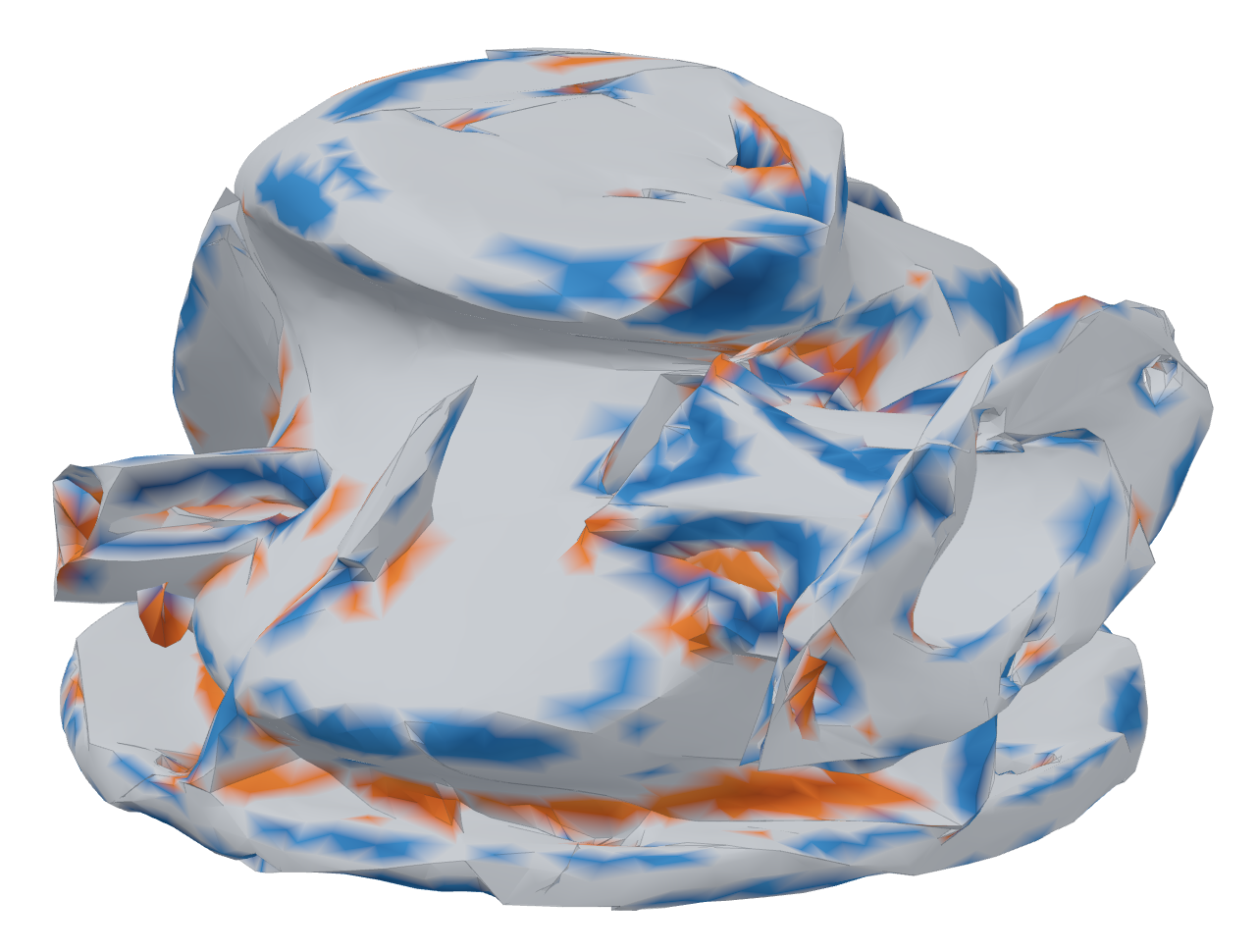}};
\node[anchor=north west,inner sep=0pt] at (177.750,157.266) {\includegraphics[width=56.250bp]{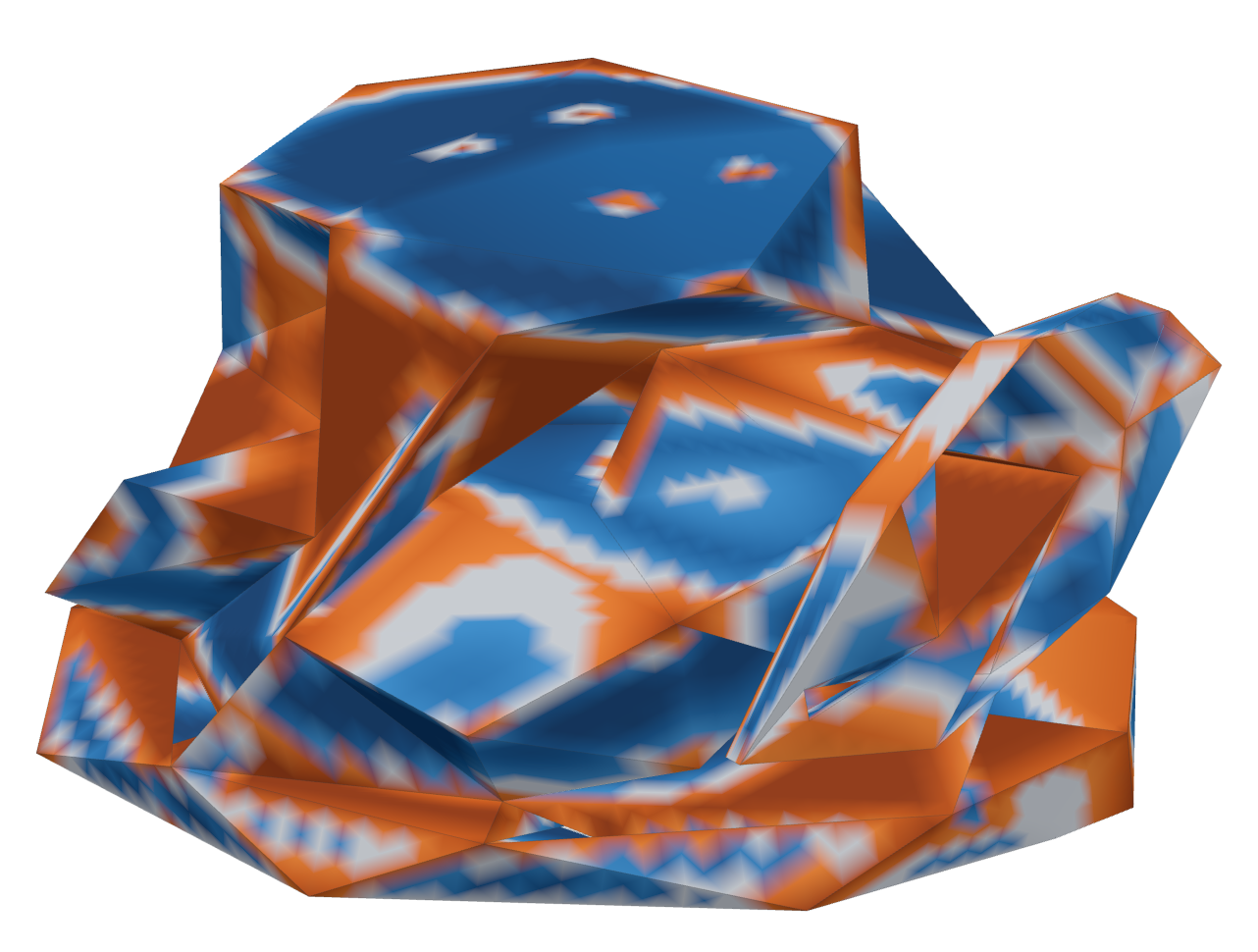}};
\node[anchor=center, inner sep=0pt, text=mercorink, font=\fontsize{7}{8.5}\selectfont] at (87.375,203.826) {91.3\%};
\node[anchor=center, inner sep=0pt, text=mercorink, font=\fontsize{7}{8.5}\selectfont] at (146.625,203.826) {62.4\%};
\node[anchor=center, inner sep=0pt, text=mercorink, font=\fontsize{7}{8.5}\selectfont] at (205.875,203.826) {10.6\%};
\node[anchor=east, inner sep=0pt, text=mercorink, font=\fontsize{7}{8.5}\selectfont] at (58.000,217.826) {Missing};
\node[anchor=west, inner sep=0pt, text=mercorink, font=\fontsize{7}{8.5}\selectfont] at (176.000,217.826) {Excess};
\node[anchor=north west,inner sep=0pt] at (62.000,214.826) {\includegraphics[width=110.000bp,height=6.000bp]{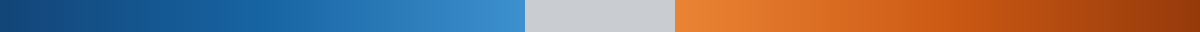}};
\draw[pumpslate,line width=0.4bp] (62.000,220.826) -- (62.000,223.326);
\node[anchor=center, inner sep=0pt, text=mercorink, font=\fontsize{6.5}{8}\selectfont] at (62.000,227.826) {$-2$\,mm};
\draw[pumpslate,line width=0.4bp] (117.000,220.826) -- (117.000,223.326);
\node[anchor=center, inner sep=0pt, text=mercorink, font=\fontsize{6.5}{8}\selectfont] at (117.000,227.826) {$0$};
\draw[pumpslate,line width=0.4bp] (172.000,220.826) -- (172.000,223.326);
\node[anchor=center, inner sep=0pt, text=mercorink, font=\fontsize{6.5}{8}\selectfont] at (172.000,227.826) {$+2$\,mm};
\end{tikzpicture}
\endgroup
\caption{\textbf{Parametric CAD reconstruction on hidden parts.} The ground truth for each of the three hidden parts, in their positive-twist variants, and the best of eight attempts by Claude Fable 5.2, DeepSeek V4.1 Flash and Beam 1.0 (trials 7, 4 and 3), selected by overall task reward. Numbers give part scores.}
\label{fig:artifacts}

%% file: figures/teaser_dev_leaderboard.tex
\centering
\begin{minipage}{0.86\textwidth}
\includegraphics[width=\linewidth]{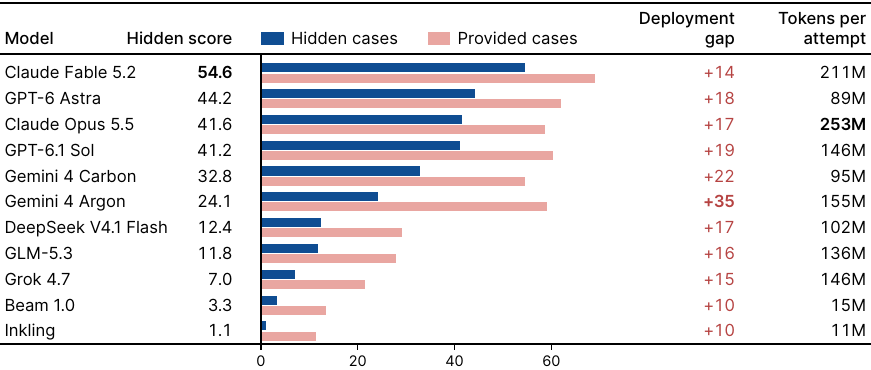}
\captionof{figure}{\textbf{Graded on the cases agents could see, every model would score higher.} Mean normalized score over all \numtasks{} tasks, for eleven models, ordered by hidden score. The dark blue bar is the score on hidden cases, and the light red bar below it is longer by the deployment gap, how much higher the model scores on the provided cases. Five tasks are not scored on provided cases and do not contribute to the gap, which is measured on the other 21. The gap is 10 to 22 points for every model except Gemini 4 Argon, whose 35-point gap would lift it from eighth to fifth among all fourteen models (\cref{fig:dev-leaderboard-all}). Tokens are input plus output tokens per attempt. The highest value in each column is in bold. The full leaderboard is in \cref{fig:leaderboard}.}
\label{fig:dev-leaderboard}
\end{minipage}

%% file: sections/principles.tex
\section{Design Principles}
\label{sec:principles}

\begin{figure}[!t]
  \centering
  \input{figures/overview_schematic_column}
  \caption{\textbf{Overview of \bench.} An agent improves a starting artifact within a time budget. An isolated verifier scores the frozen submission on development and hidden cases; their score difference is the deployment gap (development minus hidden). The agent chooses what to submit (1), anticipates transfer (2), and decides when to stop (3)}
\label{fig:overview}
\end{figure}
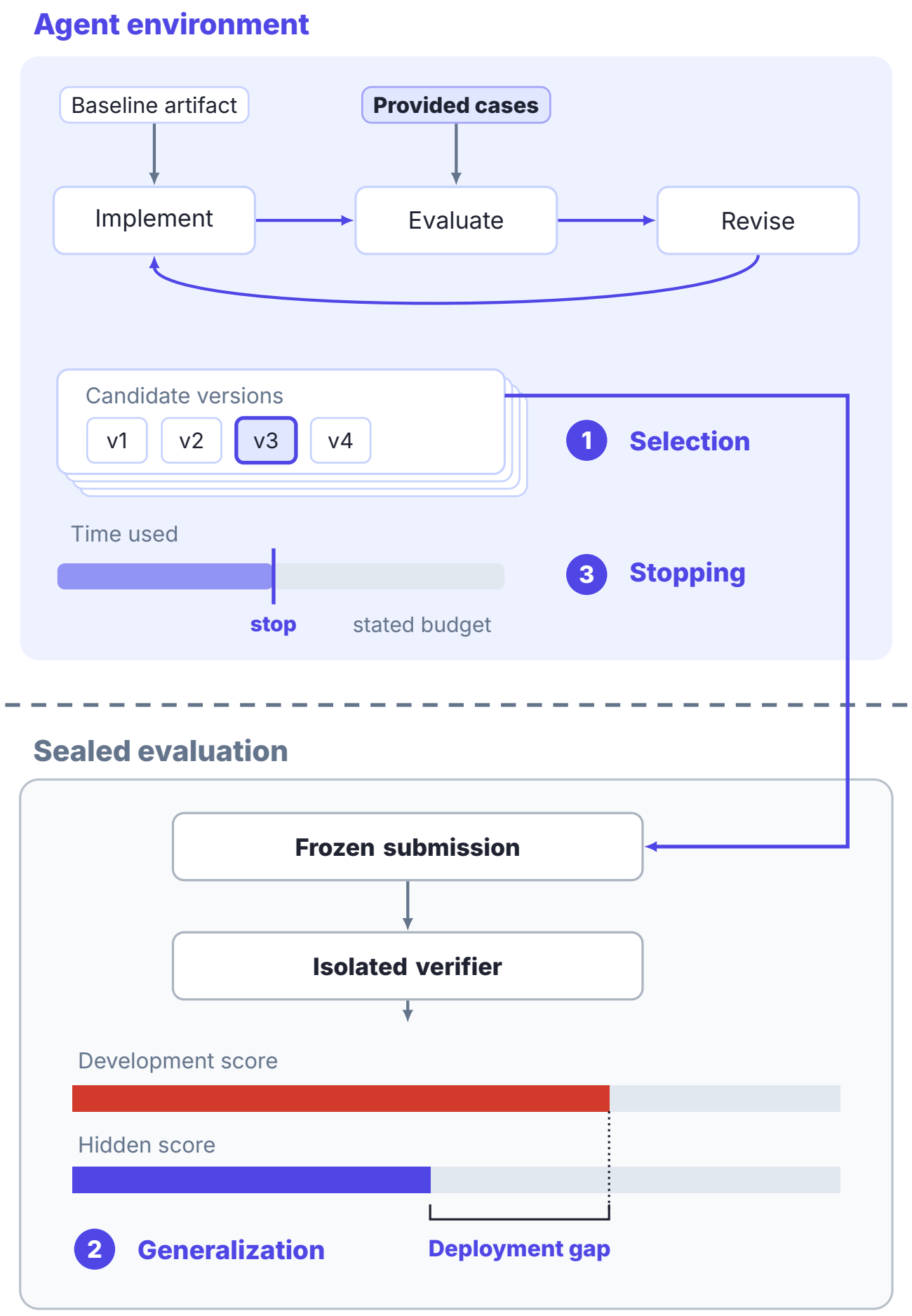

Three principles govern what agents can use, how their solutions are scored, and when they stop.

\paragraph{Information boundary.}
Each task gives the agent a starting solution, documentation, and data, examples or tests to guide development. The final solution is scored on held-out cases or operating conditions. These may follow the development distribution or differ from it to test transfer \citep{koh2021wilds}. At submission, we freeze only the declared artifact paths. An isolated offline verifier evaluates them with pristine task-supplied checking code where applicable; packages installed during development do not carry over. Scores come from execution, geometric comparison or simulation, never from an LLM judge, whose verdicts can favor their own outputs \citep{panickssery2024selfpref,pombal2026rubricspb}, miss code errors \citep{crupi2025judgecode} and be swayed by inserted text \citep{raina2024judgerobust,zhao2025onetoken}.

\paragraph{No pass mark.}
A solution can pass correctness checks and still have room to improve. Tasks therefore use continuous scores on measures such as speed, cost, geometric error and forecast skill. Each task defines a \textbf{floor} for missing, invalid or rule-breaking submissions, a \textbf{baseline} for the starting point, and an evidence-based \textbf{ceiling} that estimates attainable performance with current technology.

For the leaderboard, a task reward $r$ is normalized as $\max(0,\min(1,(r-b)/(c-b)))$, where $b$ and $c$ are its baseline and ceiling; thus the baseline maps to 0\% and the ceiling to 100\%, as in the human-normalized Atari score and RE-Bench's normalized score \citep{mnih2015human,wijk2024rebench}. Task-specific anchors are available in the \href{https://huggingface.co/datasets/mercor/OpenWorldEngineer}{dataset}.

Ceilings differ in kind. Six are exact answers, such as every chess puzzle solved; six are estimated engineering bounds, such as roofline-style limits for the GPU kernels \citep{williams2009roofline}, where reward 1 means closing 95\% of the gap to the bound; four add a margin to the best known result, such as a concrete cost 0.37\% below the best-known design; eight come from a model or reference solution of what is attainable, as for most forecasting tasks; and two are targets set by the task authors. 

Some ceilings may be out of reach, while others may be exceeded. Normalized scores therefore compare remaining room for improvement only roughly and are not percentages of human capability. There is no pass mark that tells the agent its work is finished. The agent must make that judgment, which we test in \cref{sec:split}.

\paragraph{Open book, fixed toolbox.}
Agents may use public methods, but must build with the tools and files already in the environment. They can search and read pages from a web snapshot dated \Tnowval{} \citep{exa2026snapshot}, but cannot download packages, code, data or models. Tasks are accepted only when adapting public work leaves substantial engineering (\cref{sec:frontier}). The shared snapshot and tools standardize access across evaluation dates, but do not control prior knowledge or search rankings (\cref{app:audits}).

%% file: figures/overview_schematic_column.tex
\definecolor{blockexcess}{HTML}{D23A2C}
\begin{tikzpicture}[
  x=1pt, y=1pt,
  font=\sffamily\fontsize{7}{8.4}\selectfont, text=mercorink,
  every node/.append style={execute at begin node={\hyphenpenalty=10000\exhyphenpenalty=10000}},
  chip/.style={draw=mercorline, fill=white, rounded corners=2.5pt, line width=0.5pt,
    inner xsep=3pt, inner ysep=2.5pt, font=\sffamily\fontsize{6.4}{7.4}\selectfont},
  keychip/.style={chip, fill=mercorlilac, draw=mercorindigo!60, font=\sffamily\bfseries\fontsize{6.4}{7.4}\selectfont},
  stage/.style={draw=mercorline, fill=white, rounded corners=3pt, line width=0.6pt,
    minimum height=18pt, text width=50pt, inner sep=2pt, align=center},
  sbox/.style={draw=mercorslate!55, fill=white, rounded corners=3pt, line width=0.6pt,
    minimum height=18pt, text width=120pt, inner sep=3pt, align=center, font=\sffamily\bfseries\fontsize{7}{8.4}\selectfont},
  ver/.style={draw=mercorline, fill=white, rounded corners=2pt, line width=0.5pt,
    minimum width=16pt, minimum height=12pt, inner sep=0pt, font=\sffamily\fontsize{6.3}{7}\selectfont},
  flow/.style={-{Latex[length=3.5pt,width=3pt]}, draw=mercordeep, line width=0.8pt},
  gflow/.style={-{Latex[length=3.5pt,width=3pt]}, draw=mercorslate, line width=0.8pt},
  badge/.style={circle, fill=mercordeep, text=white, inner sep=0pt, minimum size=11pt,
    font=\sffamily\bfseries\fontsize{7}{7}\selectfont},
  dtitle/.style={text=mercordeep, font=\sffamily\bfseries\fontsize{7.6}{8.8}\selectfont, anchor=west},
  small/.style={text=mercorslate, font=\sffamily\fontsize{6.4}{7.4}\selectfont},
  region/.style={anchor=west, font=\sffamily\bfseries\fontsize{8.5}{10}\selectfont},
]
\node[region, text=mercordeep] at (0,8) {Agent environment};
\fill[mercormist, rounded corners=5pt] (0,0) rectangle (234,-162);

\node[chip] (base) at (36,-13) {Baseline artifact};
\node[keychip] (dev) at (117,-13) {Provided cases};
\node[stage] (impl) at (36,-44) {Implement};
\node[stage] (eval) at (117,-44) {Evaluate};
\node[stage] (rev)  at (198,-44) {Revise};
\draw[gflow] (base.south) -- (impl.north);
\draw[gflow] (dev.south) -- (eval.north);
\draw[flow] (impl.east) -- (eval.west);
\draw[flow] (eval.east) -- (rev.west);
\draw[flow] (rev.south) .. controls (198,-70) and (36,-70) .. (impl.south);

\foreach \d in {6,4,2} \fill[white, draw=mercorline, line width=0.5pt, rounded corners=3pt] (10+\d,-84-\d) rectangle (130+\d,-112-\d);
\fill[white, rounded corners=3pt] (10,-84) rectangle (130,-112);
\draw[mercorline, line width=0.6pt, rounded corners=3pt] (10,-84) rectangle (130,-112);
\node[small, anchor=west] at (14,-91) {Candidate versions};
\node[ver] at (26,-103) {v1};
\node[ver] at (46,-103) {v2};
\node[ver, draw=mercordeep, line width=1pt, fill=mercorlilac] at (66,-103) {v3};
\node[ver] at (86,-103) {v4};
\node[badge] at (152,-103) {1};
\node[dtitle] at (160,-103) {Selection};

\node[small, anchor=west] at (10,-128) {Time used};
\fill[mercorrule, rounded corners=2pt] (10,-136) rectangle (130,-143);
\fill[mercorindigo!70, rounded corners=2pt] (10,-136) rectangle (68,-143);
\draw[mercordeep, line width=1pt] (68,-132) -- (68,-147);
\node[small, text=mercordeep, font=\sffamily\bfseries\fontsize{6.2}{7}\selectfont] at (68,-153) {stop};
\node[small, anchor=east, font=\sffamily\fontsize{6.2}{7}\selectfont] at (130,-153) {stated budget};
\node[badge] at (152,-139) {3};
\node[dtitle] at (160,-139) {Stopping};

\draw[mercorslate, line width=1pt, dash pattern=on 4pt off 3pt] (-4,-174) -- (238,-174);

\node[region, text=mercorslate] at (0,-186) {Sealed evaluation};
\fill[mercorpaper, rounded corners=5pt] (0,-194) rectangle (234,-336);
\draw[mercorslate!45, rounded corners=5pt, line width=0.6pt] (0,-194) rectangle (234,-336);
\node[sbox] (art) at (104,-212) {Frozen submission};
\node[sbox] (ver) at (104,-244) {Isolated verifier};
\draw[gflow] (art.south) -- (ver.north);
\draw[gflow] (ver.south) -- ([yshift=-6pt]ver.south);

\def\xa{14}\def\xb{220}
\node[small, anchor=west] at (\xa-2,-270) {Development score};
\fill[mercorrule] (\xa,-276) rectangle (\xb,-283);
\fill[blockexcess] (\xa,-276) rectangle (158,-283);
\node[small, anchor=west] at (\xa-2,-292) {Hidden score};
\fill[mercorrule] (\xa,-298) rectangle (\xb,-305);
\fill[mercordeep] (\xa,-298) rectangle (110,-305);
\draw[mercorink, line width=0.6pt, densely dotted] (158,-283) -- (158,-310);
\draw[mercorink, line width=0.6pt] (110,-308) -- (110,-312) -- (158,-312) -- (158,-308);
\node[anchor=north, font=\sffamily\bfseries\fontsize{6.6}{7.5}\selectfont, text=mercordeep] at (134,-314) {Deployment gap};
\node[badge] at (20,-320) {2};
\node[dtitle] at (28,-320) {Generalization};

\draw[flow, line width=1pt] (130,-91) -- (222,-91) |- (art.east);
\end{tikzpicture}

%% file: sections/benchmark.tex
\section{Tasks}
\label{sec:benchmark}

\subsection{Tasks and artifacts}
\label{sec:domains}
\bench{} contains 26 tasks across four domains: eight software engineering tasks, five physical engineering tasks, six machine learning tasks and seven scientific computing tasks. Agents produce executable programs, parametric designs and other solutions, scored on measures of correctness and performance. We provide the full task list, time budgets and scoring anchors in our \href{https://huggingface.co/datasets/mercor/OpenWorldEngineer}{Hugging Face dataset}. 

Agents have 4 to 18 hours per task (median 8) and one pass over all tasks permits 257 agent-hours. Seven tasks provide an H100 or B300 GPU. Tasks are packaged in the Harbor format \citep{harbor2026} and run on Modal \citep{modal}.

\paragraph{Software and systems.}
Eight tasks ask agents to make existing software faster or smaller while preserving required functionality. Examples include OxiLite, a SPARQL store on SQLite \citep{w3c2013sparql,gaffney2022sqlite}, a Parquet lakehouse \citep{armbrust2021lakehouse}, an exactly-once streaming pipeline under fault injection, and the MicroPython VM and LVGL renderer on an RV32 guest \citep{micropython,lvgl,waterman2019riscv}. Agents tune on public workloads. The verifier checks functional correctness on hidden use cases before scoring efficiency through measures such as execution time, size, or instruction count.

\paragraph{Models and kernels.}
Machine learning tasks ask agents to post-train or extend small language models within a compute budget, such as teaching Qwen3-8B \citep{yang2025qwen3} chess tactics or extending SmolLM2 \citep{allal2025smollm2} from 8k to 64k context. Other tasks ask agents to extract vibration features that work on unseen machines or implement attention kernels \citep{dao2022flashattention,shah2024flashattention3} on H100 and B300 GPUs. Trained models are evaluated on held-out inputs, and kernels must produce correct outputs on held-out shapes before speed counts toward the score.

\paragraph{Scientific computing.}
These tasks ask agents to implement GPU kernels for quantum-chemistry integrals \citep{li2025gpu4pyscf} and seismic wave propagation, fill cloud gaps in Sentinel-2 time series \citep{drusch2012sentinel2}, and relocate an earthquake sequence. Other tasks use past records to forecast pollutant plumes, storm surge and reaction kinetics under new conditions. Forecasts and reconstructions are scored against held-out observations or ground truth.

\paragraph{Physical engineering: CAD reconstruction.}
Three tasks ask agents to recover editable geometry from undimensioned multi-view drawings (drawing-to-CAD), noisy scans (scan-to-CAD), or calibrated drawings with packaging variants (parametric CAD). The drawing and scan tasks score reconstructions of held-out parts on geometric agreement and task-specific checks. A visually plausible output can still omit a feature or misestimate a dimension. \Cref{fig:artifacts} compares the three held-out parts in the parametric CAD task with the best reconstructions by Claude Fable 5.2, DeepSeek V4.1 Flash and Beam 1.0.

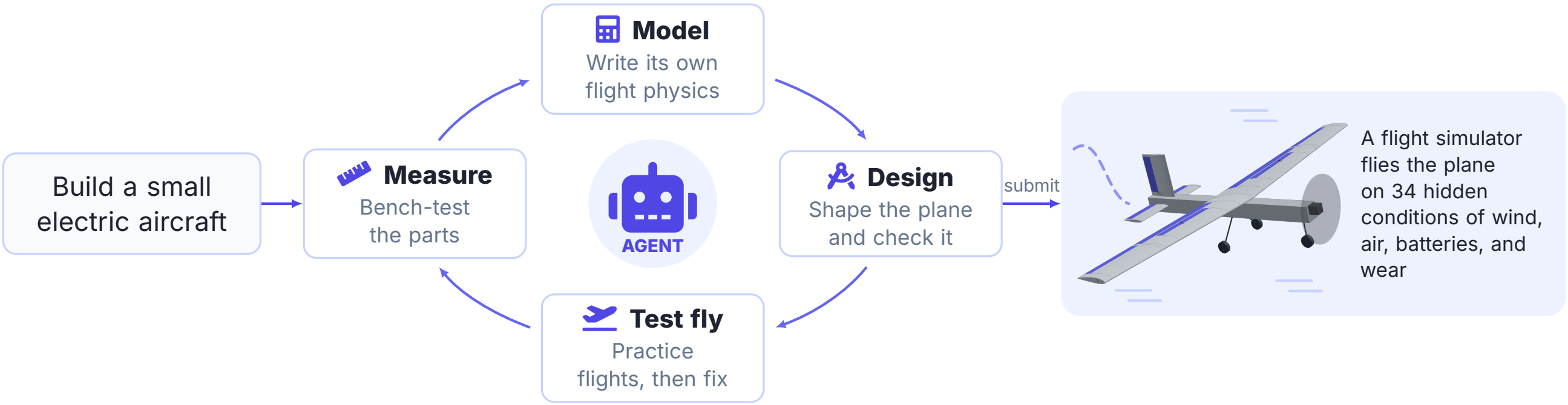
\begin{figure*}[t]
  \centering
  \input{figures/aircraft_walkthrough}
  \caption{\textbf{Aircraft design requires iterative engineering without access to the final product for evaluation.} The agent receives a brief, a working starting design, a parts catalog and the autopilot's code, but cannot run the simulator that grades it. For up to eight hours it iterates: measuring parts with budgeted bench tools, writing its own models of the physics, designing and checking the aircraft, and flying a few practice flights. A private simulator then flies the submitted aircraft in 34 hidden conditions. The aircraft shown is Claude Fable 5.2's best design, the only one of 112 attempts to reach the ceiling.}
  \label{fig:aircraft}
\end{figure*}

\paragraph{Physical engineering: Structures and aircraft.}
The concrete-building task asks agents to improve a sizing procedure for ten held-out buildings, with 6--20 storeys and varied geometry, materials, occupancy and wind exposure. A design earns a cost-based score only after passing the structural and detailing checks. The aircraft task asks agents to design a parametric electric aircraft and set its operating parameters. A simulator then evaluates the design over 34 hidden autopilot flights spanning three missions and varied wind, build and mechanism conditions. \Cref{fig:aircraft} illustrates the agent's design and testing process, from initial brief to evaluation in the simulator.

\subsection{Task difficulty and remaining headroom}
\label{sec:frontier}
We assess task difficulty using three sources of evidence: the performance of existing public methods, the targets used to define task ceilings, and the gains from giving agents more time to complete the work.

\paragraph{Public methods fall short of task targets.} Where applicable, authors ran the closest public tools end to end, including EPA AERMOD for plume dispersion \citep{cimorelli2005aermod}, VELEST and hypoDD for earthquake relocation \citep{kissling1994velest,waldhauser2000hypodd,waldhauser2001hypodd}, and ANUGA for storm surge \citep{nielsen2005anuga}. None of the tested methods scored above 0.259 against a ceiling of 1 (task versions differ from the leaderboard), so each task demands substantial engineering beyond the best public starting point.

\paragraph{Ceilings estimate achievable performance.} Targets are exact answers, engineering bounds or estimates of attainable performance, so improving on a known implementation need not saturate a task (\cref{sec:principles}).

\paragraph{Additional time leaves substantial headroom.}Across five tasks, doubling the time from half the limit to the full limit raises Opus's and Sol's scores by 10 points, yet their full-length runs still finish 50 points below the ceiling. On two of these tasks, one more hour in a new run only adds half a point (\cref{sec:split}).

%% file: figures/aircraft_walkthrough.tex
\begin{tikzpicture}[
  x=1pt, y=1pt,
  font=\sffamily, text=mercorink,
  every node/.append style={execute at begin node={\hyphenpenalty=10000\exhyphenpenalty=10000}},
  stage/.style={draw=mercorline, line width=0.6pt, rounded corners=4pt, fill=white,
    text width=62pt, inner sep=3.5pt, align=center, font=\sffamily\fontsize{7.2}{8.6}\selectfont},
  flow/.style={-{Latex[length=3.5pt,width=3pt]}, draw=mercordeep, line width=0.8pt},
  cyc/.style={-{Latex[length=3.2pt,width=2.8pt]}, draw=mercorindigo, line width=0.8pt},
]
  \newcommand{\stagetext}[3]{{\fontsize{8.5}{10}\selectfont\bfseries{\color{mercordeep}#1}\enspace #2\par}%
    \vspace{1pt}{\color{mercorslate}#3}}

  \node[draw=mercorline, line width=0.6pt, rounded corners=4pt, fill=mercorpaper, text width=66pt,
        inner sep=7pt, align=center, font=\sffamily\fontsize{9}{11}\selectfont] (task) at (40,-50)
    {Build a small electric aircraft};

  \def\cx{202}\def\cy{-50}\def\rx{74}\def\ry{45}
  \fill[mercormist] (\cx,\cy) circle (20pt);
  \node[text=mercordeep, font=\fontsize{22}{22}\selectfont] at (\cx,\cy+2) {\faRobot};
  \node[font=\sffamily\bfseries\fontsize{6}{7}\selectfont, text=mercordeep] at (\cx,\cy-13) {AGENT};
  \foreach \a/\b in {154/121, 59/26, -26/-59, -121/-154}
    \draw[cyc] ({\cx+\rx*cos(\a)},{\cy+\ry*sin(\a)}) arc[start angle=\a, end angle=\b, x radius=\rx, y radius=\ry];
  \node[stage] (measure) at (\cx-\rx,\cy)  {\stagetext{\faRuler}{Measure}{Bench-test the parts}};
  \node[stage] (model)   at (\cx,\cy+\ry)  {\stagetext{\faCalculator}{Model}{Write its own flight physics}};
  \node[stage] (design)  at (\cx+\rx,\cy)  {\stagetext{\faDraftingCompass}{Design}{Shape the plane and check it}};
  \node[stage] (fly)     at (\cx,\cy-\ry)  {\stagetext{\faPlaneDeparture}{Test fly}{Practice flights, then fix}};
  \draw[flow] (task.east) -- (measure.west);

  \fill[mercormist, rounded corners=6pt] (329,-15) rectangle (486,-85);
  \draw[flow] (design.east) -- node[above, font=\sffamily\fontsize{6}{7}\selectfont, text=mercorslate] {submit}
    (329,-50);
  \draw[mercorindigo!70, line width=0.9pt, dash pattern=on 3pt off 2.2pt, line cap=round]
    (333,-33) .. controls (341,-27) and (343,-48) .. (350,-50);
  \foreach \y/\x in {-21/382, -77/346, -74/396} {
    \draw[mercorline, line width=0.8pt, line cap=round] (\x,\y) -- ++(11,0);
    \draw[mercorline, line width=0.8pt, line cap=round] (\x+4,\y-3.2) -- ++(10,0);
  }
  \node[anchor=center, inner sep=0pt] (plane) at (376,-50)
    {\includegraphics[width=84pt]{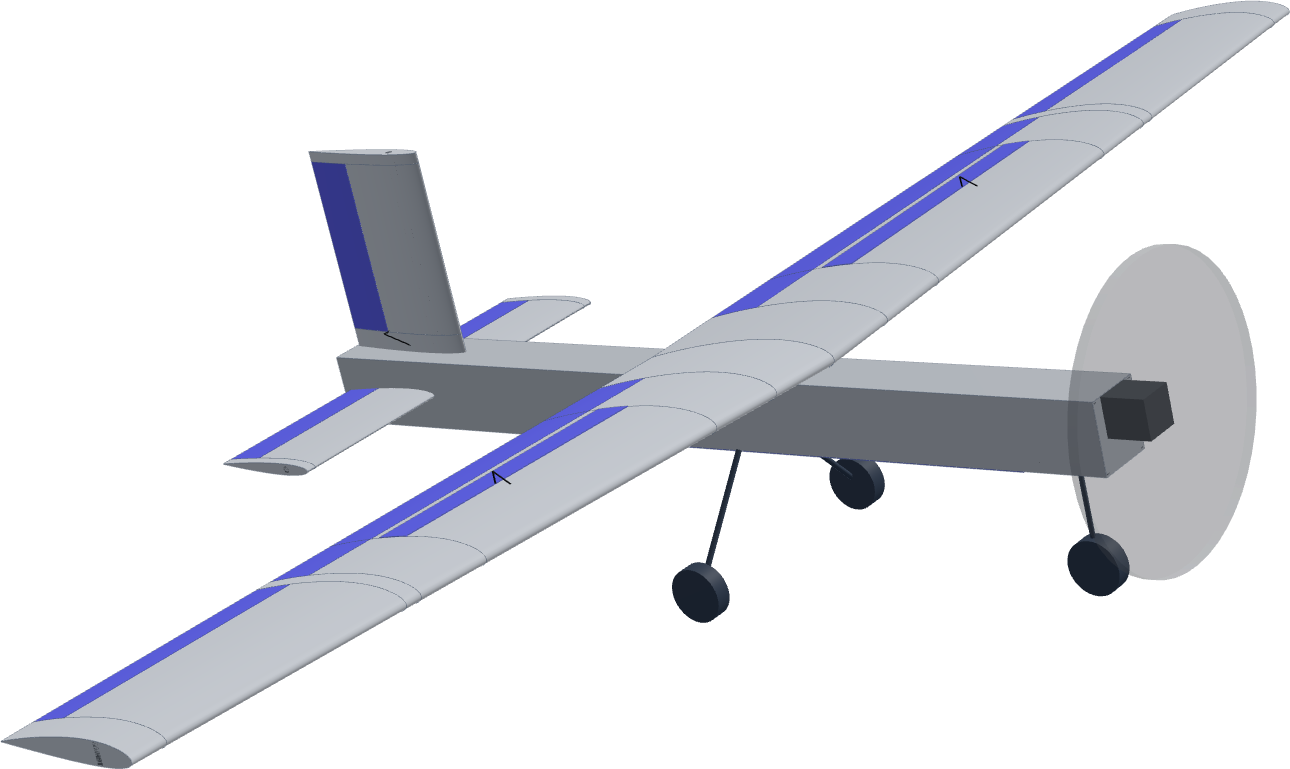}};
  \node[anchor=west, text width=60pt, align=left, inner sep=0pt,
        font=\sffamily\fontsize{6.8}{8.2}\selectfont] at (422,-50)
    {A flight simulator flies the plane on 34 hidden conditions of wind, air, batteries, and wear};
\end{tikzpicture}

%% file: sections/construction.tex
\section{How the Tasks Were Built}
\label{sec:construction}

\paragraph{Experts.}
Tasks were built by domain experts recruited through Mercor who held a master's or PhD in their field, had at least ten years of experience, or both. Before implementation, each expert proposed a task identifying relevant public work, the engineering work that remains after adapting it, the required deliverable, and how its correctness and performance would be verified. Maintainers reviewed each proposal before the expert built the task. 

\paragraph{Review gates.}
Tasks were reviewed for grading correctness, resistance to exploits and difficulty.
\begin{itemize}
  \item \textbf{Isolated grading.} The verifier runs separately from the agent's workspace; build-time checks verify that missing or invalid submissions receive the score floor and that the unchanged starting solution receives the expected baseline score on the specified hardware.
  \item \textbf{Red-team trials.} On each task, agents were explicitly instructed to maximize their scores without completing the intended engineering work, including when given access to the verifier's source code.
  \item \textbf{Difficulty trials.} Tasks were  rejected in their current form when a frontier agent's mean score over three attempts reached two-thirds of the way from the baseline to the performance ceiling.
\end{itemize}
Automated static checks and an LLM-assisted rubric review additionally assessed task structure, reproducibility, instructions, score design and time and compute budgets. Each task required approval from three human reviewers before acceptance. Roughly 400 tasks made it through automated checks, but only the 26 best were accepted to the benchmark. After evaluation, we audited every agent trajectory for unfair failures and attempted exploits (\cref{sec:exploits}).

The difficulty gate deliberately selects against tasks on which frontier agents already perform well. Low scores are therefore partly a consequence of benchmark construction, as in other adversarially filtered evaluations \citep{zellers2018swag,phan2025hle}. Our results characterize agents' capabilities and limitations on this selected set of tasks, rather than their performance on typical engineering work. Difficulty filtering can also bias comparisons against models used to select tasks \citep{phang2022adversarial}. To mitigate this risk, we draw on Phang et al.’s suggestion to use diverse adversary models, calibrating each task’s difficulty using GPT-6.1 Sol, GPT-6 Astra, Opus 5.5, and Fable 5.1.

%% file: sections/results.tex
\section{Results}
\label{sec:results}

\subsection{Protocol and model coverage}
\label{sec:protocol}
Fourteen models each ran eight independent trials per task, \LBscheduled{} trials in all (\cref{tab:pass8-summary}; \citealp{anthropic2026fable52,openai2026gpt6astra,anthropic2026opus55,openai2026gpt61sol,anthropic2026sonnet55,google2026gemini4carbon,anthropic2026fable51,google2026gemini4argon,moonshot2026kimik31,deepseek2026v41flash,deepseek2026v41flash1017,zai2026glm53,xai2026grok47,reflection2026beamtechreport,thinkingmachines2026inkling}). The label Kimi K3.1 denotes Kimi K3.1 Pro, and DeepSeek V4.1 Flash denotes the 1017 checkpoint; Gemini 4 Carbon used the model identifier \texttt{gemini-4-carbon}. Agents ran in a fork of the Proximus harness from FrontierSWE \citep{frontierswe}, with artifact submission, image viewing, and search and page fetch over the \Tnowval{} snapshot. All other network access was disabled, and the default prompt informed the agent that search was available. A model's score averages each task's eight trials and weights tasks equally. An invalid artifact is an agent failure and receives the floor. We report approximate intervals using $\pm1.96$ standard errors across task means (\citealp{miller2024errorbars}). A two-level bootstrap resampling tasks and trials changes these intervals by at most 0.3 points, because scores vary far more across tasks than across trials.

Additionally, trials in which a model intentionally cheated score zero when the exploit reached the graded artifact. PostTrainBench similarly adjusts results for cheating, giving runs flagged by its LLM judge the base model's score \citep{rank2026posttrainbench}. This changes the score of eight trials (the others already scored zero); their graded scores accompany the released audit data. Keeping those scores would change no rank.

\subsection{Leaderboard and domain results}
\label{sec:results-main}
No model is close to the ceilings, and one model leads clearly. Claude Fable 5.2 has the highest overall mean, \MeanFableTwo\% (approximate 95\% interval \LowFableTwo--\HighFableTwo\%), \LeadGap{} points ahead of GPT-6 Astra at \MeanAstra\% (\cref{tab:pass8-summary,fig:leaderboard}). Compared task by task, its lead is \LeadGapLow{} to \LeadGapHigh{} points (95\% interval), and it scores higher on \LeadTasks{} of the \numtasks{} tasks. Even the highest mean leaves \TopGap{} percentage points to the normalized ceilings. Astra, Claude Opus 5.5 (\MeanOpus\%) and GPT-6.1 Sol (\MeanSol\%) are within \NextThreeSpread{} points, and their intervals overlap widely. Claude Sonnet 5.5 follows at \MeanSonnet\%, Gemini 4 Carbon at \MeanCarbon\% and Claude Fable 5.1 at \MeanFable\%.

\paragraph{Mean and best of eight.}
Selecting each task's best graded trial adds \BestGainMin{} to \BestGainMax{} points for the seven strongest models, calculated before rounding (\cref{tab:pass8-summary}). Claude Fable 5.2 has the highest best-of-eight score, \BestFableTwo\%, ahead of Opus (\BestOpus\%) and Astra (\BestAstra\%). Best-of-eight uses the grader to choose after the fact, as pass@$k$ does \citep{kulal2019spoc,chen2021codex,brown2024monkeys}, so it shows what repeated attempts can reach. \Cref{sec:analysis} shows that an agent choosing by its own development scores keeps about half of this gain. Time use also separates the models: on average Astra and Sol each use \TimeSol\% of the time limit, Fable 5.1 \TimeFable\% and Inkling \TimeInkling\%. Few trials are cut off: 91\% end with the agent's own final submission, so time use is largely the agent's choice, and \cref{sec:split} shows that for some models this choice follows the limit they are given.

\input{figures/results_table}

\textbf{Claude Fable 5.2} leads all four domains, by \DomMarginMin{} to \DomMarginMax{} points, calculated before rounding (\cref{tab:domain-results}). Second place differs by domain: \textbf{Claude Opus 5.5} in software engineering (\DomSecondSWE\%), \textbf{GPT-6 Astra} in physical engineering (\DomSecondPE\%) and scientific computing (\DomSecondSC\%), and \textbf{Claude Sonnet 5.5} in machine learning (\DomSecondML\%). Scientific computing has the lowest leading mean.

\subsection{Token and cost efficiency}
\label{sec:efficiency}
\Cref{fig:efficiency} compares performance with the tokens and model cost consumed per attempt \citep{kapoor2024agentsmatter,kapoor2025hal}. \textbf{Claude Fable 5.2} has the highest mean while using \TokensFableTwo{} million tokens and \$\CostFableTwo{} per attempt, and it lies on both frontiers. \textbf{GPT-6 Astra} averages \MeanAstra\% with \TokensAstra{} million tokens and \$\CostAstra; \textbf{GPT-6.1 Sol} averages \MeanSol\% at \$\CostSol, and \textbf{Claude Sonnet 5.5} \MeanSonnet\% at \$\CostSonnet. Astra uses fewer total tokens than Sol, Sonnet or Fable 5.2 but costs more, illustrating why token volume and dollar cost give different comparisons. \textbf{Claude Fable 5.1} costs \$\CostFable{} per attempt for \MeanFable\%: more than Sonnet, for a lower score. 

Measured by tokens rather than time (\cref{fig:time-scaling}, left), GPT-6 Astra scores highest at every token count its runs reach, and Fable 5.2 pulls ahead only by using more tokens than Astra uses at any limit. By cost (right), GPT-6.1 Sol scores highest from its cheapest runs, about \$0.33 per attempt, up to its full-limit cost of about \$50, and DeepSeek V4.1 Flash scores highest below that. DeepSeek's full-limit runs cost about what Sol's do at 1/32 of the limit, \$1.4 per attempt, but score 12.4 against Sol's 14.7. \Cref{sec:split} demonstrates how models spend their time.

These costs exclude hardware. Each of the seven GPU tasks gives the agent one GPU for its whole run, so GPU time tracks working time: over its 56 trials on these tasks, a model used between 51 GPU-hours (Inkling) and 714 (Carbon) of the 740 available, and Astra, Sol and Carbon used more than 90\%.

%% file: figures/results_table.tex
\begin{table}[t]
  \caption{\textbf{Mean and best-of-eight scores, with time and cost.} Scores in \%, normalized and averaged with tasks weighted equally. \emph{Best}: best of eight, each task's highest graded trial. \emph{At 0\%}: tasks on which no trial beat the baseline. \emph{Time}: share of the agent time limit used, averaged per attempt. \emph{Cost}: model-usage USD per attempt at list prices.}
  \label{tab:pass8-summary}
  \centering\ResearchTableStyle\fontsize{7.7}{10.5}\selectfont\setlength{\tabcolsep}{2.4pt}%
  \begin{tabular}{@{}l r c r r r r@{}}
    \toprule
    \headrow \textbf{Model} & \textbf{Mean} & \textbf{95\% CI} & \textbf{Best} & \textbf{At 0\%} & \textbf{Time} & \textbf{Cost} \\
    \midrule
    Claude Fable 5.2 & 54.6 & 46.2--63.0 & 68.3 & 0 & 86 & 105.08 \\
    GPT-6 Astra & 44.2 & 36.5--51.9 & 58.3 & 0 & 95 & 234.63 \\
    Claude Opus 5.5 & 41.6 & 36.0--47.1 & 58.4 & 0 & 75 & 75.70 \\
    GPT-6.1 Sol & 41.2 & 34.3--48.0 & 57.1 & 0 & 95 & 52.12 \\
    Claude Sonnet 5.5 & 39.0 & 30.8--47.2 & 52.7 & 1 & 82 & 38.65 \\
    Gemini 4 Carbon & 32.8 & 27.2--38.3 & 51.0 & 0 & 92 & 195.51 \\
    Claude Fable 5.1 & 29.1 & 24.1--34.1 & 45.6 & 0 & 65 & 87.70 \\
    Gemini 4 Argon & 24.1 & 18.3--29.9 & 42.2 & 1 & 78 & 49.44 \\
    Kimi K3.1 & 13.6 & 9.7--17.4 & 20.6 & 1 & 61 & 21.16 \\
    DeepSeek V4.1 Flash & 12.4 & 9.3--15.4 & 23.4 & 1 & 56 & 1.37 \\
    GLM-5.3 & 11.8 & 8.4--15.2 & 21.0 & 1 & 74 & 24.83 \\
    Grok 4.7 & 7.0 & 4.4--9.6 & 13.8 & 3 & 63 & 146.28 \\
    Beam 1.0 & 3.3 & 2.1--4.5 & 8.9 & 3 & 32 & 1.47 \\
    Inkling & 1.1 & 0.5--1.8 & 3.8 & 6 & 8 & 3.26 \\
    \bottomrule
  \end{tabular}
\end{table}

%% file: sections/analysis.tex
\section{Can Agents Tell Whether Their Work Will Hold Up?}
\label{sec:analysis}

An agent never sees the hidden evaluation cases, so how it judges its own work rests on its internal knowledge, provided examples, held-out splits of its own data or tests, and anything accessible on the web. This evidence serves two distinct purposes: determining which of its versions is better, and whether a version will hold up in deployment. We expect the first to be easier than the second, much as new test sets lower image classifiers' accuracy but largely keep their ranking \citep{recht2019imagenet,miller2021accuracy}. To measure both, we score each final artifact on its task's provided cases with the same formula as the hidden grader, so every trial gets a development score next to its hidden score. This is also the score that a benchmark grading agents on the cases they could see would report. It measures the evidence an agent has to judge by, not the agent's own judgment, and on some tasks, such as aircraft design, the agent cannot compute it directly. To measure the judgment itself, we asked each model to forecast the hidden and development scores of each of its runs from the run's log (\cref{app:selfest}). We leave out the four forecasting tasks because their artifact is itself the forecast for the hidden cases, and the B300 attention engine because grading its development workloads would be too costly (\cref{app:dev}).

\begin{table}[t]
  \caption{\textbf{Development evidence by domain.} Normalized points, averaged over models, with tasks weighted equally. \emph{Gain}: hidden score of the attempt with the best development score minus that of a random attempt. \emph{Share}: gain divided by the best attempt's (oracle's) score minus random. \emph{Gap}: mean development minus mean hidden score. \Cref{tab:dev-results-full} gives each task, with its class, scores and intervals.}
  \label{tab:dev-results}
  \centering\ResearchTableStyle
  \input{figures/dev_results_table_domain}
\end{table}

\begin{figure*}[t]
  \centering
  \includegraphics[width=\textwidth]{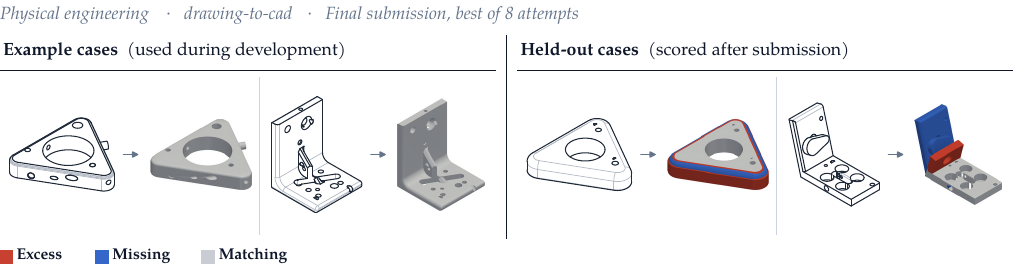}
 \caption{\textbf{The deployment gap in drawing-to-CAD: Gemini 4 Argon rebuilds the example cases exactly but not similar held-out cases.} Each drawing (isometric view shown) is paired with the part built from it by Argon's best attempt, coloured against the true part as \textcolor[HTML]{C8402F}{\rule{0.65em}{0.65em}} excess, \textcolor[HTML]{3566C9}{\rule{0.65em}{0.65em}} missing or \textcolor[HTML]{C9CDD3}{\rule{0.65em}{0.65em}} matching material. Both example parts score 1.00, but on held-out cases of the same kinds of part the attempt misses details the examples did not exercise: an oversized outline and cut-back top edge on the flange (score 0.14) and a short, straight block in place of the bracket's leaning upright (0.00).}
  \label{fig:argon-d2c}
\end{figure*}

\subsection{Agents can tell which of their versions is better}
\label{sec:selection}
Development scores tell an agent which of its attempts is likely to do better on the hidden cases. For each model and task, we pick one of the eight attempts by its development score, much as code generators pick among samples by running tests \citep{li2022alphacode,chen2022codet}, and compare its hidden score with that of a random attempt and of the oracle, the attempt the hidden grader would pick in hindsight. Among eight attempts, picking by development score raises the hidden score by 6.4 points over a random attempt (95\% CI 5.1 to 7.6), about half of the oracle's gain (\cref{fig:dev-selection,tab:dev-results}). Selecting based on development scores doesn’t depend on any hidden scoring, so an agent can use it (unlike the oracle, which requires knowing the hidden scores). The gain is positive on 20 of the 21 tasks (its 95\% interval is above zero on 13), does not hinge on any one model or task, and replicates on new attempts. It also holds on new cases that no agent saw: on fresh aircraft flights and concrete buildings, picking by development score still clearly beats a random attempt (\cref{app:dev-methods}). Ranked by development scores, the fourteen models also keep most of their hidden order (rank correlation 0.96). The agents' own forecasts rank their attempts too: picking by forecast recovers 41\% of the oracle's gain, against 50\% for the development score on the same attempts (\cref{app:selfest}).

Agents also make the same judgment within a run, as they must choose which artifact they want graded (each submission is a pinned commit, and the last one is graded). On average, the artifact the agent chooses to submit only falls short of the best artifact by under 2 points and picking among a run's submissions by development score would gain less than a point over the agent's own choice. This suggests that agents are already effective at judging what their best work looks like (\cref{app:within-run}).

\subsection{They cannot tell whether a version will hold up beyond what they tested}
\label{sec:dev-levels}
Agents expect their work to hold up far better than it does. Asked to forecast the hidden score of each of their runs, all fourteen models overestimated it (\cref{app:selfest}), by 21 points on average (95\% CI 18 to 25). Their 80\% intervals held the hidden score in only 55\% of runs and in 44\% it fell below them.

Development scores are a poor guide to how well a version will hold up. Final artifacts score 18 points higher on provided cases than on hidden ones (95\% CI 16.0 to 19.3; \cref{tab:dev-results}), and 12 points higher without the two tasks whose development data come from another domain by design (\cref{app:dev-models}). A benchmark that graded agents on the cases they could see would report these higher scores: on our tasks, it would put the three strongest models we scored this way at 64 instead of 48 (\cref{fig:dev-leaderboard-all}). Agents expected a gap of just 5.5 points (95\% CI 4.6–6.3), underestimating the actual gap by roughly two-thirds. The results suggest that agents' miscalibration is not simply due to inaccurate estimates of development performance, but rather showcases a failure in anticipating how their solutions will perform on unseen conditions. Even on the 14 tasks where agents closely predicted their development scores, they expected only a 7-point gap compared to the observed 20-point gap (\cref{app:selfest}).

The gap depends on what the hidden cases test (\cref{tab:dev-results-full}). Averaged by class, it is 10.6 points where the provided cases are drawn like the hidden ones, 20.9 where they come with their answers and 23.8 where they differ by design, or 9.0 without the two transfer tasks. The agents expected about a fifth of it where the cases are drawn alike and a quarter where they differ by design, but about half where the provided cases come with their answers, where our pre-registered test narrowly fails (\cref{app:selfest}). On five of the nine tasks drawn like the hidden ones (chess puzzles, decode attention, seismic simulation, draft-head training and the concrete buildings), development scores are at most 3.1 points higher than the hidden scores, similar to the gap between Kaggle’s public and private leaderboards \citep{roelofs2019metaanalysis}. Where the hidden cases test something the provided cases do not, the gap grows and the ranking fails with it: across tasks, the more development scores overstate, the less of the oracle's gain they recover (rank correlation $-0.81$ across the 21 tasks). Bearing-fault transfer and Sentinel-2 gap filling ask for models that work on new machines or new regions. Their development scores run 67 to 70 points high, and on bearing transfer every artifact scores perfectly on its development data, so the score cannot even rank attempts. Agents expected a gap here, but of only 15 to 18 points, and forecast hidden scores of 75 to 82 against actual 24 to 33. On software tasks, hidden workloads catch failures that public ones miss, even where the two are drawn alike, as stronger test suites do for generated code \citep{liu2023evalplus,ahmed2025overfitting}. One software attempt in seven passes every development check yet scores zero on the hidden workloads. As a result, selecting by development score recovers only a third of the oracle's gain, illustrating how false positives limit resampling with imperfect verifiers \citep{stroebl2024inferencescaling}. Where provided cases ship with their answers, scores run high but mostly keep their order: on drawing-to-CAD and parametric CAD, and on aircraft design, whose test flights differ by design, development scores run 13 to 19 points high, yet picking by them still recovers 66 to 83\% of the oracle's gain.

The exception is one model whose development scores run far ahead of its hidden ones. Gemini 4 Argon's development scores exceed its hidden ones by 35 points, about twice the median of the other thirteen models. Going only by development score would lift Argon from eighth to fifth (\cref{fig:dev-leaderboard-all}), and its self-estimates of performance overshoot the most out of any model, by 40 points. Argon ranks first on the development parts of drawing-to-CAD and scan-to-CAD, and seventh and eighth on the hidden parts (\cref{app:dev-models}), and on the three CAD tasks its attempt with the best development score does no better than a random one.

\subsection{More checking does not close the gap: the checks look like the cases agents have}
\label{sec:within-run}
\label{sec:selfcheck}
If agents misjudged their work because they did not check it, more checking would help. We examined this through pipeline inspection, check removal, and an experiment encouraging additional validation. We see that agents already perform many checks. An audit of all runs found the agent's own estimate of its score in 89\% of them. Similar to the agents' post-run forecasts, these in-run estimates overstate hidden performance by 25 points. Even when agents knew their development scores, they overestimated hidden performance by 18.5 points, about as much as the 20.4-point gap between development and hidden scores on those tasks. This suggests the checks agents build for themselves provide little more information than the provided cases do. Half the runs name a risk that the work will not generalize, yet those runs' forecasts overshoot as much as the rest's (\cref{app:selfest}). In six models' final pipelines, 92\% of drawing- and scan-to-CAD pipelines compare the solid they built with the input, as in visual self-verification for CAD code \citep{alrashedy2025cadcodeverify}. Yet pipelines that check have about the same gap as those that do not, Argon checks in 14 of 16 pipelines and has the largest gaps, and removing the checks from Sol's pipelines lowers its development scores more than its hidden ones (\cref{app:selfcheck}). Telling agents to validate does not help either. We gave finished full-length runs on the concrete and page layout tasks one more hour to work, with either a neutral instruction or one to check each change against the current submission on the provided cases and keep the original if nothing validates as better. With the validation instruction, agents re-submitted their artifact unchanged about seven times as often (18\% against 3\%), yet scored no better ($-0.1$ points, 95\% CI $-1.8$ to $1.7$; \cref{app:continuation}). In nearly nine of ten continuations that ended worse than they started, the development score held or rose.

These findings suggest that the limitation is not simply how often agents check their work, but what their checks cover. Where development scores misled, the failed attempts show agents failing to account for cases outside of those they were given (\cref{app:dev-why}). GPT-6.1 Sol sped up MicroPython's practice programs rather than the interpreter and judged its speed only on them and Argon added shortcuts to the SPARQL store that relied on quirks of the practice data. Additional checks against the same limited cases may reinforce apparent improvements without revealing failures under new conditions.

\section{Can Agents Tell When to Stop?}
\label{sec:split}
A time limit can constrain performance by ending a run, but it can also influence how an agent plans and allocates effort before the limit is reached. Which of the two matters depends on the model. We ran six models at varying time limits, from 1/128 to 1/2 of each task's limit, and compared them with their leaderboard runs at the full limit (\cref{fig:time-scaling}; \cref{app:time-sweep}). The two GPT models work right up to the limit. GPT-6 Astra and GPT-6.1 Sol use essentially all of every limit and make use of background commands so they can keep working while waiting for results. The two Claude models stop on their own. Claude Fable 5.2 uses 82 to 88\% of whatever limit it is given and Claude Opus 5.5 60 to 75\%, across a 128-fold range. They start background jobs less often, and at the longer limits they spend about half their time waiting for commands to finish. They also submit about twice as many versions as the GPT models. Working to the limit pays when time is short: Astra leads Fable 5.2 by 5 to 7 points at every limit up to 1/8, and Fable 5.2 beats it on at most 4 of the 26 tasks at 1/16 and below. But the Claude models gain more as the limit grows. For the GPT models each doubling adds about the same, roughly 5 points, while for the Claude models the gain rises from about 2 points at the shortest limits to about 10 at the longest. Fable 5.2 draws level with Astra at a quarter of the limit and leads by 10 points at the full one, and Opus trails both GPT models at every shorter limit but draws level with them at the full limit. GLM-5.3 uses most of each limit and DeepSeek V4.1 Flash even less as the limit grows, but each gains only about 11 points across the whole range. Because the Claude models consistently use a set percentage of whatever limit they are allocated, it appears they plan their work around the provided constraints rather than the inherent demands of the task.

We isolate this effect by varying Claude Opus 5.5's stated time budget while keeping the task and tools fixed. When told it had eight hours instead of four, Opus scored 14.3 points higher on aircraft design (95\% CI 8.2 to 20.7) and 7.1 points higher on storm-surge forecasting (3.6 to 10.5),\footnote{The main text uses normalized points, as on the leaderboard, except for the search study, which reports task reward. The tests were pre-registered on task reward, which \cref{app:split} reports; it gives the same conclusions.} even though most of its eight-hour runs finished within four hours (38 and 40 of 48; \cref{fig:time-limit}). Under the four-hour limit, Opus stopped sooner, often after roughly an hour on aircraft design, leaving most of its allotted time unused. No run reached its limit, so the only difference was the limit Opus was told. Opus reads its deadline early in a run and paces its work to fit. Across the five tasks of the split test below, halving the leaderboard's limit cost it 11 points, and on the aircraft task, raising the limit from eight hours to sixteen added another 10.8 (2.1 to 19.3). Other models behave differently. On the aircraft task, GLM-5.3 worked the same hours and scored the same whether it was told it had eight hours or sixteen (\cref{app:split}). More time spent working does not always improve the final submission: on the aircraft task, some agents kept working past their best design and ended up submitting a worse one (\cref{app:within-run}). The stated time limit is therefore part of the evaluation condition, not merely a cap on available compute. It can change performance without being reached and, because models respond to it differently, it can change the gaps between them.

Because agents can tell which attempt is better (\cref{sec:selection}), one might trade time for attempts and let development scores choose. Stopping early costs more than choosing wins back. Earlier work found that the better split depends on the setting: in RE-Bench some agents did best with many 30-minute attempts and others with fewer 2-hour ones \citep{wijk2024rebench}, and on math problems the best mix of sequential and parallel compute depends on difficulty \citep{snell2024scaling}. In a randomized test on five tasks that span all four domains, Opus and Sol were run once with the full time limit or twice with half the limit each, submitting whichever scored better on the provided cases (\cref{app:split}). The full-length run wins by 5.7 points (95\% CI 4.2 to 7.2). A single half-length run scores 9.7 points below the full-length run, and picking the better of two by development score wins back 4.0 of them. Even picking the better one in hindsight loses by 3.4 points. On no task did two shorter runs beat one longer one. The margin follows what the lost time is worth: 13 points on model training, 7 to 8 on the building and aircraft designs, and a tie on page layout and gap-filling, where half the time costs least. Returns do flatten toward the end of a run: on the concrete and page layout tasks, the hidden score rises by about 9 points in the second quarter of a full-length run and 2 in the last, and one more hour in a new run adds half a point (95\% CI $-1.0$ to $2.0$; \cref{app:within-run,app:continuation}).  Therefore, although the marginal gains from longer runs diminish, they remain more valuable than the additional selection opportunities created by splitting the budget. 

%% file: figures/dev_results_table_domain.tex
\begin{tabular}{@{}l r r r r@{}}
\toprule
\headrow \textbf{Domain} & \textbf{Tasks} & \textbf{Gain} & \textbf{Share} & \textbf{Gap} \\
\midrule
Physical engineering & 5 & $+$10.4 & 71\% & $+$13.0 \\
Machine learning & 5 & $+$6.9 & 71\% & $+$15.2 \\
Scientific computing & 3 & $+$2.9 & 48\% & $+$22.4 \\
Software engineering & 8 & $+$4.9 & 34\% & $+$20.1 \\
\midrule
\textbf{All tasks} & 21 & $+$6.4 & 52\% & $+$17.6 \\
\bottomrule
\end{tabular}

%% file: sections/cheating.tex
\section{Reward Hacking and Cheating}
\label{sec:exploits}

An agent that cannot see how it will be graded may try to reach the grader instead. Long, open-ended runs give an agent many chances to game its grader \citep{metr2025rewardhacking,baker2025monitoring,bondarenko2025specgaming,zhong2025impossiblebench}, so we checked every trial. Exploit attempts are rare, no attempt to tamper with the grader raised a score, and no exploit changes a model's rank. We count a trial as an \emph{exploit attempt} when the agent's commands show it pursuing the score without doing the intended engineering work, for example by altering evaluation machinery, recovering hidden answers or exploiting gaps between the task and its checks. LLM reviewers read every trial, and each counted finding was checked by hand (\cref{app:exploits}).

Across the fourteen models, 38 of \LBscheduled{} trials (1.3\%) contain an attempt. Tampering with the grader never raised a score, because the verifier uses its own pristine source tree and checking code; Inkling, for example, edited a repository's patch collector to hide changes outside the allowed paths. Attempts to recover hidden answers also failed: Sol brute-forced private generator seeds without finding them. Attempts that exploited the task itself earned little. Argon's hard-coded fast paths for public MicroPython functions failed hidden checks, and parametric CAD programs that returned stored public models scored little above the task's baseline, which is the score of such a copy. Eight attempts earned a nonzero reward; we set those scores to zero, which lowers no model's mean by more than 0.13 points. Attempts are not confined to the weakest models: Argon, in the middle of the leaderboard, has the highest attempt rate, 5.3\% (\cref{app:exploit-attempts}).

%% file: sections/discussion.tex
\section{Discussion}
\label{sec:discussion}

\paragraph{Telling when the job is done.}
Our findings distinguish an agent's ability to improve engineering work from its ability to judge when that work is complete. Agents often select their strongest submitted versions, but development performance can substantially overstate performance on unseen cases, the agents themselves expect far higher scores than they get, and stated deadlines can influence when agents stop even without being reached. Benchmarks with pass/fail grading measure whether a submission meets a set of criteria, while benchmarks with a final grader allow agents to optimize directly against the specific evaluation conditions. Neither establishes how well an agent can judge its work without final evaluation feedback. Graded on the cases they could see, every model in our study would score higher, and the model that fit those cases most would gain most (Argon would rank fifth rather than eighth). Separating development from evaluation therefore changes both the level of an agent benchmark's results and, where an agent fits its provided cases, the ordering of models.

\paragraph{What development evidence is good for.}
Running several attempts and submitting the one with the best development score recovers about half of the best-of-eight gain without a hidden grade, and within a run, agents already submit their best version most of the time; retrospectively selecting among their submissions offers little additional gain, while instructing agents to validate against existing provided cases does not improve hidden performance in our follow-up experiments. Choosing well does not replace time, though, because the limit shapes how much an agent does: in our randomized test, one long run beat two shorter ones on three of five tasks and tied on the other two, so time is better spent on longer runs than on more attempts. Development scores stop being a guide where an agent has fit its provided cases, as Argon did on the CAD tasks and three software tasks. A choice is only as trustworthy as the evidence the agent had no incentive to game. Why some agents fit their provided cases is open. One speculative explanation, which we have not tested for any model, is training that rewards passing the tests an agent can see. Developers have reported models learning to special-case or bypass unit tests during reinforcement learning on coding tasks \citep{anthropic2025claude37card,baker2025monitoring}.

\paragraph{Instructions change agent behavior more than their scores, except the time limit.} Instructing agents on how to work changed what the agents did, but not how well they scored. Told to research prior work, nearly every agent searched (\cref{app:followup}) and told to validate, agents held back their changes (\cref{sec:within-run}). The time limit changed both: Opus did less and scored lower with half the limit, and did more and scored higher with twice the limit, without reaching either (\cref{sec:split}). With no pass mark to stop at (\cref{sec:principles}), how much of the limit an agent uses is largely its own choice, and models differ in whether they plan around the limit at all: the Claude models stop at a set share of whatever limit they are told, while the GPT models work right up to it (\cref{sec:split}). A benchmark's time limit is therefore part of its task: scores at one limit need not carry over to another, the order of the models changes with it (\cref{fig:time-scaling}), and time used belongs next to each score, as in \cref{tab:pass8-summary}.

\paragraph{Access to published methods is not the bottleneck.}
\label{sec:ablation}
\label{sec:prompts}
Offering web search does not measurably change scores, even when nearly every agent searches. Five models ran every task with and without search over the snapshot, and search changed their mean task reward by \SearchDelta{} (95\% CI \SearchCI{}). Under the default prompt only \SearchUptake{} of trials used it, so we also told Claude Opus 5.5 and GPT-6.1 Sol to research prior work before writing any code. On five tasks, 79 of 80 such trials searched, yet they scored the same as trials given the same instruction without web access (task reward difference \MatchedPooledDelta{}, 95\% CI \MatchedPooledCI{}). The trajectories offer a possible explanation. Agents often name the relevant method before they search. Search mostly supplies facts about the local setup, such as an instruction's syntax or a library's API, and ideas the same model reaches without it: 132 of the 180 retrieved ideas that reached a final artifact also appear in that model's offline runs. The hard part of these tasks is adapting known methods to the problem at hand, not finding them (\cref{app:search}).

\paragraph{Limitations.}
Tasks were accepted partly because agents struggled with them, so the scores describe this suite rather than research and engineering work in general. The tasks are synthetic worlds built by experts, and simulated performance is not a substitute for real life. \numtasks{} tasks give five to eight per domain. Provided cases differ in kind. Some differ from the hidden cases by design, as in sentinel2-gapfill and bearing-fault-transfer, so a development--hidden gap there is expected and is not overfitting; \cref{tab:dev-results-full} gives each task's class.  Agents forecast their scores after the run, from its log, rather than at submission. Five tasks have no development evaluation. The validation test hands a finished artifact to a new run rather than extending the original one. The split test compares one run with two half-length runs at each task's own time limit---more and shorter runs were not tested, and longer limits were tested only for Opus on two tasks and GLM-5.3 on one. Scores depend on each task's metric, baseline and estimated ceiling. The search results apply to these tasks, agents and retrieval tools (\cref{app:search}).

\subsection{Release}
\label{sec:release}
We release all tasks and evaluation trajectories with the leaderboard. Every task file and released trajectory carries a canary string so that they can be filtered from training data \citep{srivastava2023bigbench}, and evaluations with search must use a snapshot dated before the tasks' release, so that these materials cannot be retrieved \citep{han2025stc}.

\section{Conclusion}
\label{sec:conclusion}
\bench{} asks whether agents can tell when the job is done: whether their work will hold up on cases they never saw, and when it is good enough to stop.  Across \numtasks{} tasks, agents often submit their strongest versions, and development evidence helps select better solutions across independent attempts. But a solution that performs well on the cases used to develop it may fail on new workloads, inputs or conditions. Across 21 tasks, final artifacts scored 18 points higher on provided cases than on unseen cases, and asked to forecast their scores on the unseen cases, agents overestimated them by 21 points. Additional validation against the same limited evidence did not reliably uncover these failures. Stated time limits also shaped when agents stopped, even when those limits were never reached. Together, these findings expose a gap between improving an engineering solution and knowing whether it is ready to deliver. As agents take on greater responsibility for engineering work, the challenge is not only to improve the solutions they produce, but also their ability to judge what the available evidence establishes, what remains uncertain, and when further work is justified.

%% file: sections/appendix_tasks.tex
\section{Task List}
\label{app:tasks}

\begin{table*}[p]
  \caption{All \numtasks{} tasks and their scoring anchors. \emph{Hours}: agent time limit. Floor and baseline are task rewards before leaderboard normalization, rounded to three decimals, from the same frozen scoring version as \cref{fig:leaderboard}; every ceiling is a task reward of 1, and \cref{tab:ceilings} says what it means for each task. The leaderboard maps baseline to 0\% and ceiling to 100\%, clipping lower rewards to 0\%. Parametric CAD's baseline is the score of copying the public reference of the requested part class.}
  \label{tab:tasks}
  \centering\ResearchTableStyle\scriptsize\renewcommand{\arraystretch}{1.1}%
  \setlength{\tabcolsep}{3pt}
  \begin{tabular}{@{}>{\raggedright\arraybackslash}p{0.26\textwidth} >{\raggedright\arraybackslash}p{0.45\textwidth} r c r r@{}}
    \toprule
    \headrow \textbf{Task} & \textbf{Objective} & \textbf{Hours} & \textbf{GPU} & \textbf{Floor} & \textbf{Baseline} \\
    \midrule
    \multicolumn{6}{@{}l}{\textit{Software engineering}} \\
    \nolinkurl{function-page-layout} & Order functions \citep{pettis1990profile} so a binary touches fewer pages on unseen startup scenarios & 4 & -- & 0.000 & 0.000 \\
    \nolinkurl{lvgl-speed} & Cut RV32 guest instructions for LVGL's software renderer \citep{lvgl} on hidden scenes, framebuffers byte-identical & 16 & -- & 0.000 & 0.000 \\
    \nolinkurl{micropython-speed} & Cut RV32 guest instructions for MicroPython's VM \citep{micropython} on hidden programs, exact outputs, no runtime codegen & 16 & -- & 0.000 & 0.000 \\
    \nolinkurl{oxilite-speed} & Make a SPARQL \citep{w3c2013sparql} store on SQLite \citep{gaffney2022sqlite} load faster, store smaller and query with less work, answers unchanged & 16 & -- & 0.100 & 0.100 \\
    \nolinkurl{skew-join-engine} & Exact grouped cyclic joins under skew, correlation and changing filters & 4 & -- & 0.000 & 0.003 \\
    \nolinkurl{sql-replay} & Compress a PostgreSQL change stream into a shorter SQL script that replays it & 6 & -- & 0.000 & 0.000 \\
    \nolinkurl{stateful-parquet} & Stateful incremental Parquet lakehouse under mixed mutation and analytical load \citep{apacheparquet,armbrust2021lakehouse} & 4 & -- & 0.150 & 0.163 \\
    \nolinkurl{streaming-etl-payments} & Exactly-once streaming ETL (Kafka CDC to Iceberg plus alerts; \citealp{kreps2011kafka,apacheiceberg}); maximize sustained rate under fault injection & 16.5 & -- & 0.000 & 0.000 \\
    \addlinespace
    \multicolumn{6}{@{}l}{\textit{Physical engineering}} \\
    \nolinkurl{concrete-multistorey-building} & Cheaper NBR~6118-compliant \citep{abnt2014nbr6118} reinforced-concrete frame designs for unseen buildings & 4 & -- & 0.000 & 0.000 \\
    \nolinkurl{dbf-aircraft-design} & Parametric CAD aircraft for a Design/Build/Fly mission \citep{aiaadbf}, with budgeted tests on a private simulator & 8 & -- & 0.000 & 0.227 \\
    \nolinkurl{drawing-to-cad} & Editable CadQuery programs \citep{cadquery} from undimensioned multi-view DXF drawings of machined parts & 16 & -- & 0.000 & 0.000 \\
    \nolinkurl{parametric-cad-reconstruction} & Native free-form CAD from calibrated drawings, then smooth packaging variants & 8 & -- & 0.000 & 0.098 \\
    \nolinkurl{scan-to-cad} & Editable CadQuery programs from noisy 3D scans of machined parts & 16 & -- & 0.000 & 0.000 \\
    \addlinespace
    \multicolumn{6}{@{}l}{\textit{Machine learning}} \\
    \nolinkurl{b300-attention-engine} & Speed up a pairwise-gated attention engine for training, prefill and decode & 17 & B300 & 0.000 & 0.000 \\
    \nolinkurl{bearing-fault-transfer} & Vibration representation for bearing faults that transfers to unseen machines under a fixed linear probe & 8 & -- & 0.000 & 0.000 \\
    \nolinkurl{chess-puzzle-post-training} & Post-train Qwen3-8B \citep{yang2025qwen3} on chess tactics within a fixed service budget & 17.5 & H100 & 0.000 & 0.033 \\
    \nolinkurl{int4-decode-attention} & Decode-attention kernels for a 4-bit paged KV cache \citep{kwon2023pagedattention}, four model shapes & 12 & H100 & 0.000 & 0.000 \\
    \nolinkurl{smollm2-long-context} & Extend SmolLM2-1.7B \citep{allal2025smollm2} from 8k to 64k context within a fixed compute budget & 8 & H100 & 0.000 & 0.000 \\
    \nolinkurl{smollm3-draft-head} & Train a speculative-decoding \citep{leviathan2023speculative} draft head for SmolLM3-3B \citep{smollm3} within a fixed compute budget & 8 & H100 & 0.000 & 0.000 \\
    \addlinespace
    \multicolumn{6}{@{}l}{\textit{Scientific computing}} \\
    \nolinkurl{gpu4pyscf-jk-speed} & Accelerate FP64 four-center Coulomb and exchange integral kernels in GPU4PySCF \citep{li2025gpu4pyscf} & 18 & H100 & 0.000 & 0.000 \\
    \nolinkurl{plume-dispersion-forecast} & Forecast hourly pollutant concentrations at receptors for planned stack releases from tracer and meteorological records & 6 & -- & 0.000 & 0.023 \\
    \nolinkurl{relocate-earthquake-sequence} & Relocate an earthquake sequence (epicentres and depths) from supplied seismic observations & 4 & -- & 0.000 & 0.072 \\
    \nolinkurl{seismic-wave-propagation} & Multi-step 3D acoustic wave propagator as device-only CUDA, compiled by the grader with NVRTC & 12 & H100 & 0.000 & 0.000 \\
    \nolinkurl{sentinel2-gapfill} & Reconstruct cloud gaps in Sentinel-2 time series with uncertainty, using Sentinel-1 SAR \citep{drusch2012sentinel2,torres2012sentinel1} & 4 & -- & 0.000 & 0.000 \\
    \nolinkurl{spectral-kinetics-forecast} & Forecast species concentrations over time for new operating cases of a catalysed reaction & 4 & -- & 0.000 & 0.082 \\
    \nolinkurl{storm-surge-forecast} & Forecast design-storm still-water levels, including wave setup, at coastal locations & 4 & -- & 0.000 & 0.036 \\
    \bottomrule
  \end{tabular}
\end{table*}

\begin{table*}[p]
  \caption{What a task reward of 1 means. Every task's ceiling is a reward of 1; this table states it in the task's own terms. \emph{Basis}: X, an exact answer; B, an estimated or measured engineering bound; K, the best known result plus a margin; M, a model or reference solution of attainable performance; A, a target set by the task's authors. $^\dagger$Reward is $\ln(1/(1-p))/\ln 20$ for progress $p$ from the baseline toward the bound, so each halving of the remaining gap earns 0.23 and reward 1 closes 95\% of it.}
  \label{tab:ceilings}
  \centering\ResearchTableStyle\scriptsize\renewcommand{\arraystretch}{1.1}%
  \setlength{\tabcolsep}{3pt}
  \begin{tabular}{@{}>{\raggedright\arraybackslash}p{0.26\textwidth} >{\raggedright\arraybackslash}p{0.66\textwidth} c@{}}
    \toprule
    \headrow \textbf{Task} & \textbf{Reward 1 means} & \textbf{Basis} \\
    \midrule
    \multicolumn{3}{@{}l}{\textit{Software engineering}} \\
    \nolinkurl{function-page-layout} & The worst startup family closes 40\% of the log page gap to ideal packing; the best known training-only order closes 35\% & K \\
    \nolinkurl{lvgl-speed} & 6.64$\times$ fewer guest instructions, from a cost model of an efficient renderer (model range 3.7--8.3$\times$) & M \\
    \nolinkurl{micropython-speed} & 6.47$\times$ fewer guest instructions, with every interpreter cost component at its estimated floor & M \\
    \nolinkurl{oxilite-speed} & A measured probe store plus modelled headroom: load 10$\times$ cheaper, store 4.5$\times$ smaller, queries about 6$\times$ cheaper & M \\
    \nolinkurl{skew-join-engine} & Each database reaches an operation-count bound, 5--69$\times$ faster than the starter (geometric mean 13.4$\times$) & B \\
    \nolinkurl{sql-replay} & A replay script 94\% the length of the shortest known exact replay & K \\
    \nolinkurl{stateful-parquet} & Probe-based estimates, all at once: 0.30$\times$ the starter's latency, 0.45$\times$ its memory, 0.55$\times$ its writes & M \\
    \nolinkurl{streaming-etl-payments} & Ramp step 23 (169{,}407 events/s), the first step at least 95\% of the way, counted in 1.25$\times$ ramp steps, to an estimated bound of 211{,}758 events/s & B$^\dagger$ \\
    \addlinespace
    \multicolumn{3}{@{}l}{\textit{Physical engineering}} \\
    \nolinkurl{concrete-multistorey-building} & A design 0.37\% cheaper than the best-known design, for every building & K \\
    \nolinkurl{dbf-aircraft-design} & The contest score's 100-point cap, set by hand-calibrated references for each mission & A \\
    \nolinkurl{drawing-to-cad} & Exact geometry: every hidden part within 0.005\,mm, or drawing identically to its input & X \\
    \nolinkurl{parametric-cad-reconstruction} & Exact target geometry: volume IoU, surface coverage and feature recall all 1 & X \\
    \nolinkurl{scan-to-cad} & Exact geometry: every hidden part within 0.005\,mm & X \\
    \addlinespace
    \multicolumn{3}{@{}l}{\textit{Machine learning}} \\
    \nolinkurl{b300-attention-engine} & 5.94$\times$ speedup, 95\% of the gap to an estimated bound of 6.2$\times$ over the starter & B$^\dagger$ \\
    \nolinkurl{bearing-fault-transfer} & Balanced accuracy 0.375 (chance 0.125): the best known probe, 0.36, plus a margin & K \\
    \nolinkurl{chess-puzzle-post-training} & All 1,000 private puzzles solved under both sampling seeds & X \\
    \nolinkurl{int4-decode-attention} & 95\% of the gap to a measured memory-read bound, the fastest kernel that only reads the cache & B$^\dagger$ \\
    \nolinkurl{smollm2-long-context} & All 225 private long-context questions answered exactly & X \\
    \nolinkurl{smollm3-draft-head} & 5 accepted tokens per verification step, the level EAGLE-3 reports (the format allows 8) & A \\
    \addlinespace
    \multicolumn{3}{@{}l}{\textit{Scientific computing}} \\
    \nolinkurl{gpu4pyscf-jk-speed} & 19.6$\times$ speedup, 95\% of the gap to an estimated 20.5$\times$ bound with kernels at 35\% of FP64 peak & B$^\dagger$ \\
    \nolinkurl{plume-dispersion-forecast} & The skill of a forecaster fitted to the public campaign, using the generator's model forms & M \\
    \nolinkurl{relocate-earthquake-sequence} & The accuracy of a locator in a 3D velocity model inverted from the public picks & M \\
    \nolinkurl{seismic-wave-propagation} & 95\% of the gap to an estimated attainable 56\% of roofline efficiency & B$^\dagger$ \\
    \nolinkurl{sentinel2-gapfill} & The probabilistic skill (CRPS) of an oracle trained on the hidden windows' labels & M \\
    \nolinkurl{spectral-kinetics-forecast} & Every one of the 600 concentrations forecast exactly & X \\
    \nolinkurl{storm-surge-forecast} & A forecaster with the true hydrodynamics and beach slopes estimated from public samples & M \\
    \bottomrule
  \end{tabular}
\end{table*}

\subsection{Example: chess post-training}
\label{sec:chess-example}
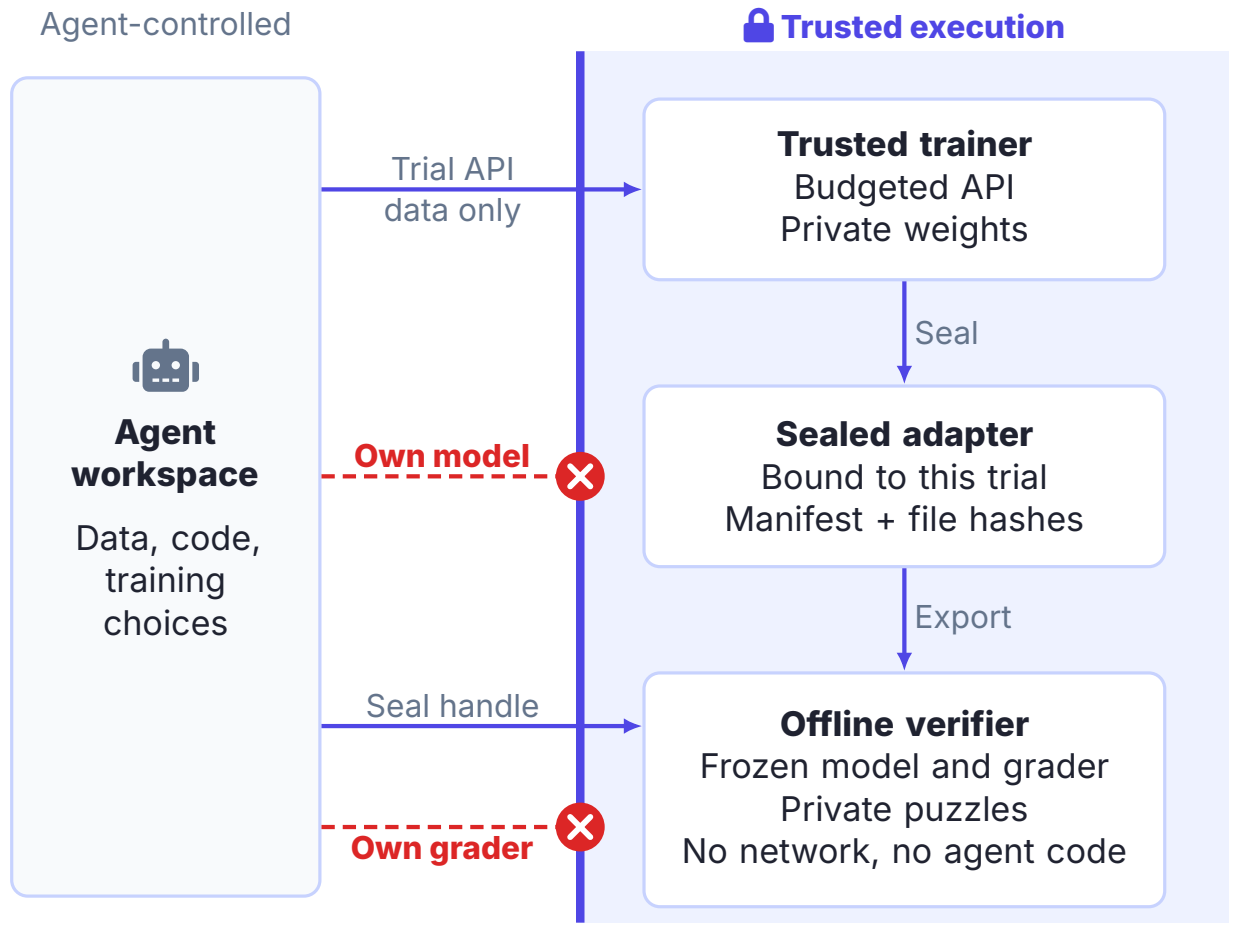
\begin{figure}[!htb]
  \centering
  \input{figures/chess_architecture}
  \caption{\textbf{Chess post-training with a sealed model handoff.} The agent improves Qwen3-8B through a training API within a budget of 17.5 hours. The vertical line marks the trust boundary. Only training data and the seal handle cross it, so the agent cannot substitute its own model or grader (red crosses). The service stores checkpoints privately and seals the selected LoRA adapter \citep{hu2022lora} with file hashes bound to the trial. After the training sandbox shuts down, a separate offline verifier checks the handle and export, then tests the model on 1,000 private puzzles under two sampling seeds. Grading executes no participant code.}
  \label{fig:chess-architecture}
\end{figure}

%% file: figures/chess_architecture.tex
\definecolor{blockred}{HTML}{DC2626}
\begin{tikzpicture}[
  x=1pt, y=1pt,
  font=\sffamily\scriptsize, text=mercorink,
  box/.style={draw=mercorline, line width=0.6pt, rounded corners=3pt,
    align=center, inner sep=4pt, minimum height=34pt},
  flow/.style={-{Latex[length=3.5pt,width=3pt]}, draw=mercordeep, line width=0.8pt},
  blocked/.style={draw=blockred, line width=0.8pt, dash pattern=on 2.5pt off 1.5pt},
  note/.style={font=\sffamily\fontsize{6.5}{8}\selectfont, text=mercorslate,
    inner sep=1.5pt},
  bad/.style={font=\sffamily\bfseries\fontsize{6.5}{8}\selectfont, text=blockred,
    inner sep=1.5pt},
  cross/.pic={
    \fill[blockred] (0,0) circle (4.6pt);
    \draw[white, line width=1.1pt, line cap=round]
      (-1.9pt,-1.9pt) -- (1.9pt,1.9pt) (-1.9pt,1.9pt) -- (1.9pt,-1.9pt);
  }
]
  \fill[mercormist] (108,26) rectangle (230,-138);
  \draw[mercordeep, line width=1.6pt] (108,26) -- (108,-138);
  \node[note, anchor=south] at (30,26) {Agent-controlled};
  \node[note, anchor=south, text=mercordeep,
    font=\sffamily\bfseries\fontsize{6.5}{8}\selectfont]
    at (169,26) {\faLock\ Trusted execution};

  \node[box, fill=mercorpaper, text width=50pt, minimum height=154pt] (agent) at (30,-56)
    {{\color{mercorslate}\normalsize\faRobot}\\[3pt]\textbf{Agent}\\\textbf{workspace}\\[4pt]Data, code,\\training\\choices};

  \node[box, fill=white, text width=90pt] (trainer) at (169,0)
    {\textbf{Trusted trainer}\\Budgeted API\\Private weights};
  \node[box, fill=white, text width=90pt] (adapter) at (169,-54)
    {\textbf{Sealed adapter}\\Bound to this trial\\Manifest + file hashes};
  \node[box, fill=white, text width=90pt, minimum height=44pt] (verifier) at (169,-113)
    {\textbf{Offline verifier}\\Frozen model and grader\\Private puzzles\\No network, no agent code};

  \draw[flow] (trainer.south) -- node[note, right] {Seal} (adapter.north);
  \draw[flow] (adapter.south) -- node[note, right] {Export} (verifier.north);

  \draw[flow] (agent.east |- 0,0) -- (trainer.west);
  \node[note, anchor=south] at (84,0) {Trial API};
  \node[note, anchor=north] at (84,0) {data only};
  \draw[flow] (agent.east |- 0,-101) -- (verifier.west |- 0,-101);
  \node[note, anchor=south] at (84,-101) {Seal handle};

  \draw[blocked] (agent.east |- 0,-54) -- (108,-54);
  \pic at (108,-54) {cross};
  \node[bad, anchor=south] at (82,-54) {Own model};
  \draw[blocked] (agent.east |- 0,-120) -- (108,-120);
  \pic at (108,-120) {cross};
  \node[bad, anchor=north] at (82,-120) {Own grader};
\end{tikzpicture}

%% file: sections/appendix_dev.tex
\section{Development-Evidence Analyses}
\label{app:dev}

\subsection{Eligibility and development sets}
\label{app:dev-eligibility}
A task is eligible only if its final artifact can be scored on the agent's provided cases with the hidden grader's formula. Twenty-one tasks are eligible, and \cref{tab:dev-results-full} gives each one's class. The four forecasting tasks are not, because their artifact is a fixed forecast or catalog for the hidden cases, and grading the B300 attention engine on its 48 development workloads takes about 1.5 B300-hours per artifact, so it was not run. Chess's development puzzles ship with their answers but are drawn like the hidden ones, so we class them by distribution. The pooled analyses cover all fourteen models; the follow-up tests below name the models they use.

\subsection{Estimands and tests}
\label{app:dev-methods}
For each model--task cell of eight attempts, three policies choose one attempt: \emph{random}, the mean of the eight hidden scores; \emph{development selection}, the hidden score of the attempt with the highest development score, with ties averaged; and the \emph{oracle}, the highest hidden score. Selection uses the raw development reward, because normalization clipping can create ties that are not real. Some task rewards are also capped inside the task's own formula. Page layout's reward is the worst startup family's mean page saving divided by 0.40, capped at 1, and bearing-fault-transfer's is clipped at a target balanced accuracy; where an artifact reaches such a cap on the provided cases, selection falls to the tie rule even if the uncapped objective still separates candidates. The primary endpoint is development selection minus random, pooled with tasks weighted equally and models equally within a task, with a task-stratified bootstrap interval and a sign-flip permutation test. The $k$-curves are exact expectations over all subsets of size $k$.

Development and random selection never use hidden grades, so their difference is unbiased on the hidden set and the oracle is optimistic, which makes the share recovered conservative. Two fresh suites test whether selection holds on new cases, in task reward units. On five new aircraft seeds (170 flights per design), development selection beats random by 0.167 (95\% CI 0.106 to 0.237), against 0.200 on the original flights. On 16 new concrete buildings, it beats random by 0.101 (0.067 to 0.136) and beats the original oracle's choice by 0.030 ($p=0.031$): the development score picks better than one hidden grading does.

On the normalized scale, choosing the better of two half-length runs by development score beats a random one by 0.040 (95\% CI 0.034 to 0.046), close to the 0.041 that the same two models' leaderboard attempts give for two attempts on these tasks. On concrete and page layout, with the rule fixed in advance, development selection among 24 new pools of eight full-length runs beats random by 0.100 (95\% CI 0.075 to 0.125; exact sign-flip $p<10^{-4}$) and recovers 55\% of the oracle's gain, against 0.106 on the same task--model cells in the main analysis. Sol's page-layout runs all reached the development cap in these pools, so that cell gains nothing there or in the main analysis.

\subsection{Pooled and per-task results}
\label{app:dev-results}
Pooled over 294 cells, a random attempt, development selection and the oracle score 25.8, 32.2 and 38.0 points. Development selection minus random is 6.4 points (95\% CI 5.1 to 7.6; $p<10^{-4}$). It is positive in 215 cells, zero in 34 and negative in 44. Leaving out one model gives 6.1 to 6.7 points, and leaving out one task 5.7 to 6.7. Of 896 software attempts, 134 pass every development gate yet score zero on hidden workloads. The gain varies less by kind of provided cases than the gap does: 6.3 points where the provided cases are drawn like the hidden ones, 6.6 where they carry their answers and 6.3 where they differ by design. On three software tasks, Argon's development pick scores zero on the hidden workloads. 

\begin{table*}[!tp]
  \caption{\textbf{Development evidence by task, with scores and intervals.} Hidden normalized score, in points, of a random attempt, the attempt with the best development score and the best attempt (oracle), averaged over model--task cells of eight attempts with models weighted equally. \emph{Gain}: development selection minus random, with a 95\% interval from bootstrapping models. \emph{Share}: gain divided by oracle minus random. \emph{Gap}: mean development minus mean hidden score. $\rho$: rank correlation between models' mean development and hidden scores; undefined for bearing, whose development score is 1.0 for every artifact. Domain and pooled rows weight tasks equally; their shares are ratios of means. Sentinel-2's gain interval has a positive lower bound (0.034 points), which rounds to 0.0.}
  \label{tab:dev-results-full}
  \centering\ResearchTableStyle\setlength{\tabcolsep}{4.5pt}%
  \input{figures/dev_results_table}
\end{table*}

\begin{figure}[!htb]
  \centering
  \includegraphics[width=\columnwidth]{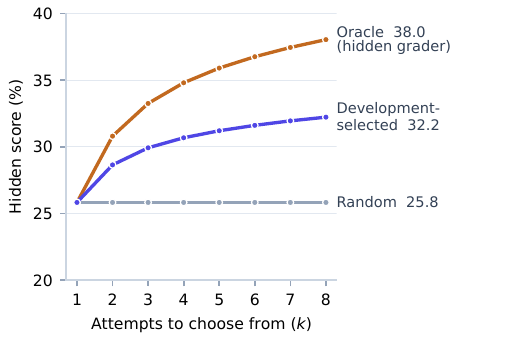}
  \caption{\textbf{Choosing by development score captures about half of the gain of choosing in hindsight.} Expected hidden score when one of $k$ attempts is submitted: a random attempt, the attempt with the best development score, or the best attempt in hindsight (the oracle). Exact expectations over all subsets of $k$ attempts, as in the unbiased pass@$k$ estimator \citep{chen2021codex}, pooled over the 21 eligible tasks with tasks and models weighted equally. The share captured is 52--57\% for every $k$ from 2 to 8. At $k=8$, development selection adds 6.4 points over a random attempt (95\% CI 5.1 to 7.6).}
  \label{fig:dev-selection}
\end{figure}

\subsection{Model-level gaps}
\label{app:dev-models}
\begin{figure}[!htb]
  \centering
  \includegraphics[width=\columnwidth]{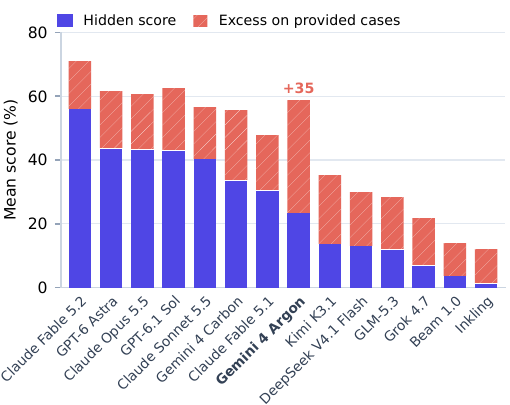}
  \caption{\textbf{Graded on the cases agents could see, Gemini 4 Argon would rise from eighth to fifth.} \Cref{fig:dev-leaderboard} for all fourteen models: mean normalized score over the 21 tasks with development scores, ordered by hidden score, with the excess of the development score stacked on the hidden score. Development scores exceed hidden ones by 10 to 22 points for thirteen models and by 35 for Argon.}
  \label{fig:dev-leaderboard-all}
\end{figure}

\Cref{fig:dev-leaderboard-all} scores each model as the leaderboard does: the mean over its eligible tasks of its mean over eight attempts, with tasks weighted equally. Intervals resample the 21 tasks. Development scores exceed hidden ones by 10.0 to 21.9 points for thirteen of the fourteen models (median 16.7) and by 35.1 for Gemini 4 Argon (95\% CI 21.6 to 49.1), 18.8 points more than the mean of the others (8.3 to 30.0). The rank correlation between models' development and hidden scores is 0.96 (0.85 to 0.99). Argon ranks fifth on provided cases and eighth on hidden ones. It rises in 98.5\% of resamples, and by three places or more in 65\%. On drawing-to-CAD and scan-to-CAD Argon ranks first on provided cases, but seventh and eighth, respectively, on hidden cases. Most of the rise comes from the CAD tasks: without them, Argon ranks seventh on provided cases and eighth on hidden ones, though its gap of 29.9 points is still the largest. Without the two transfer tasks, whose development data come from another domain by design, the mean gap falls to 12.3 points; Argon's gap is then 31.9 points, still the largest, and it still rises from eighth to fifth. Claude Fable 5.2 leads on both, with a gap of 14.5 points; the next three models, within a point of each other on hidden cases, change order, with GPT-6.1 Sol first among them on provided cases and Argon fourth.

\subsection{Controls for the gap}
\label{app:dev-controls}
A development--hidden gap can reflect answer availability, a deliberate shift, harder hidden cases, scoring differences or execution noise, so a raw gap is not evidence of overfitting. Where a fixed reference design exists, we compare the agents' gap with the reference's gap on the same split. On concrete-multistorey-building, where both splits were scored in one batch for six models, the agents' gap exceeds that of the best-known designs by 0.107 reward (95\% CI 0.019 to 0.187, resampling attempts and buildings) and about half of the excess follows building size, which weighs more in the hidden set. 

\subsection{What the failed attempts tested}
\label{app:dev-why}
Each task page carries a failure analysis that reads all eight attempts of one model and divides the points it lost among failure modes. On the seven tasks where development scores choose worst (\cref{tab:dev-results-full}), the main failures are these.
\begin{itemize}
  \item \textbf{SQL replay} (Opus): stopped about an hour early, believing little was left to gain, without testing on larger, unfamiliar workloads, and stress-tested only patterns its rules already handled.
  \item \textbf{MicroPython} (Sol): built shortcuts that sped up single practice programs instead of the core interpreter, and judged speed only on the practice programs.
  \item \textbf{Parquet} (Opus): never tested an unusual 1-million-row first batch at full size, and rewrote the whole table on every update.
  \item \textbf{OxiLite} (Argon): added shortcuts that relied on quirks of the practice data without testing them on altered queries or renamed data.
  \item \textbf{Page layout} (Opus): trusted test results too noisy to tell its layouts apart.
  \item \textbf{Bearing transfer} (Sol): fine-tuned against its own simulator, which was far easier than the new machines, and assumed that some fault patterns would persist across them.
  \item \textbf{Sentinel-2} (Opus): kept a learned network even after it did worse on a new region.
\end{itemize}
In five of the seven, the main failure is a test the agent did not run on cases like the hidden ones. Each analysis covers one model, so it shows how attempts failed rather than how often each failure occurs across models.

\subsection{Agents' own forecasts}
\label{app:selfest}
Each trial's own model, by the same provider route and at the same reasoning effort as its run, read the task instructions and the trial's log as released with the trajectories and, in one call without tools, forecast the trial's hidden reward (a median and an 80\% interval) and its development reward. The prompt named the contrast between development and hidden cases and asked for the main risk, which if anything should lower forecasts. Scores are normalized and weighted as in \cref{tab:dev-results}, intervals resample models, and selection by forecast uses the 272 model--task cells with eight forecasts. Every model's mean forecast exceeds its mean hidden score, by 13 points (Beam 1.0) to 40 (Gemini 4 Argon). The plan, fixed before any forecast was read, counted the claim that agents cannot tell whether their work will hold up as supported if forecasts exceed hidden scores and agents expect less than half of the gap where it is largest. The first condition holds; the second holds where the cases differ by design (agents expect 27\% of the gap) and narrowly fails where provided cases come with their answers (55\%). The 14 tasks of \cref{sec:dev-levels}, chosen after the results, are those on which agents' forecasts of their development scores fall within 5 points. In the audit of all runs, Claude Opus 5.5 read every log blind to hidden scores and recorded the agent's last stated expectation of its score, the last result of a check it built, its last development measurement and whether it named a risk that its work would not generalize. A run's own estimate is the first of these three numbers it has, in that order. On a 10\% sample, a second model agreed with the reader on 82 to 91\% of these items. The comparison of runs that name a risk with those that do not was not in the plan.

\subsection{Within-run selection}
\label{app:within-run}
Every submission is a pinned git commit, and the last one is graded. We graded every earlier submission on the hidden and provided cases and measured self-selection regret: the best own submission's hidden score minus the graded submission's, over trials with two or more distinct submissions.

\paragraph{Results.} Regret is 0.018 reward (95\% CI 0.015 to 0.022; 0.017 on the normalized scale). By task it is 0.047 on aircraft, 0.026 on page layout, 0.022 on parametric CAD, 0.008 on concrete and under 0.004 on drawing- and scan-to-CAD. On concrete and CAD most of it is grader noise, with the noise-adjusted regret being about 0.003. Choosing among a run's own submissions by development score improves on the graded one by 0.006 (Holm $p=0.051$).

\paragraph{Prospective test.} We repeated the analysis on the split test's 576 concrete and page-layout runs (\cref{app:split}), with a pre-registered plan, adding snapshots at 25, 50 and 75\% of each run's time to its submissions. Picking the checkpoint with the best development score changes the hidden score against the graded final by $-0.0008$ (95\% CI $-0.0040$ to $0.0023$; $p=0.63$), while the best checkpoint in hindsight beats the final by 0.028 (0.022 to 0.035), about half of it grading noise on concrete. In Sol's page-layout runs the development score sits at its cap, so the pick, with ties going to the later checkpoint, is always the final; on concrete alone the change is also $-0.0008$ ($-0.0036$ to $0.0018$). Across a run's checkpoints, development and hidden scores agree only moderately (mean rank correlation 0.63).

\subsection{Fixed-budget split test}
\label{app:split}
The test covers five tasks from all four domains: concrete-multistorey-building (4 hours), dbf-aircraft-design (8 hours), function-page-layout (4 hours), sentinel2-gapfill (4 hours) and smollm2-long-context (8 hours on one H100). Models are Claude Opus 5.5 and GPT-6.1 Sol at maximum effort, configured as on the leaderboard. Arm F is one run with budget $T$. Arm H is two independent runs with budget $T/2$ each, submitting the one with the better development score (ties go to the first). The arms differ only in the time budget, are submitted interleaved, and have 48 replicates per arm in each of the ten task--model cells, 1,440 runs in all. Each task's design was frozen before its runs were launched. The outcome is the hidden task reward. The primary endpoint is H $-$ F, with cells weighted equally, a stratified 95\% bootstrap interval and a sign-flip permutation test. The pooled analysis selects by uncapped development scores. We also report results using the capped rule originally prespecified for concrete and page layout.

\paragraph{Results.} H $-$ F is $-0.053$ (95\% CI $-0.067$ to $-0.039$; $p<10^{-5}$), and \cref{tab:split} gives each task. Halving the budget costs 0.091 (H-random $-$ F, 0.078 to 0.103), development selection recovers 0.037 of it (H $-$ H-random, 0.031 to 0.043), and H-oracle $-$ F is $-0.032$ ($-0.045$ to $-0.018$). A random half-length run scores below the full-length run in all ten cells, and the development pick scores above a random half-length run in all ten. H $-$ F is negative in nine cells, and its interval excludes zero in five, all negative. The half-length run with the higher development score is the better one on hidden cases in 69\% of pairs (95\% CI 65 to 73), but in only 50--52\% on sentinel2, whose development score also barely separates leaderboard attempts (\cref{tab:dev-results-full}). The capped rule gives $-0.056$, normalized scores $-0.057$ ($-0.072$ to $-0.042$; halving $-0.097$, selection $+0.040$, hindsight $-0.034$), and excluding replicates with a provider-degraded run $-0.053$.

\paragraph{Time use.} Opus never ran out of time: all 720 of its runs ended with its own final submission, and only one of the 716 with timing data used more than 97\% of its limit. It used the same share of its limit in both arms (full-length / half-length): 32\% / 29\% on aircraft, 84\% / 79\% on concrete, 75\% / 73\% on page layout, 73\% / 71\% on sentinel2 and 92\% / 91\% on smollm2. On aircraft, 38 of Opus's 48 full-length runs finished within four hours and its half-length runs stopped after a median 1.1 hours, yet a random half-length run scored 0.111 below a full-length run (0.063 to 0.161; normalized 0.143, 0.082 to 0.207). Sol used more than 97\% of its limit in 173 of the 240 full-length runs and 414 of the 480 half-length runs with timing data. A random half-length run scored 0.102 below a full-length run for Opus (0.109 normalized) and 0.079 for Sol. H $-$ F is $-0.066$ for Opus ($-0.087$ to $-0.046$) and $-0.040$ for Sol ($-0.059$ to $-0.021$); the difference between the models is not significant ($-0.026$, $-0.055$ to $0.002$). The arms used about the same compute: in each cell, two half-length runs together used 0.91--1.06 times the agent-hours of one full-length run and 0.80--1.23 times its model spend.

\paragraph{Why Opus stops.} We read all 720 Opus trajectories. On concrete, page layout, sentinel2 and smollm2, Opus reads its deadline in its first or second step and stops once too little time is left for another change and a full check (``With about 23 minutes left, there's not enough time for another change plus a full validation run''). The margin it keeps shrinks with the limit, but less than in proportion: half-length runs keep 60--67\% of a full-length run's median minutes, so Opus stops at a similar share of either limit. On aircraft, all 144 Opus runs used all twelve test flights and submitted a median 15 to 20 minutes after the last one, often with hours left (``I have about three hours left but no flight tests remaining, so more changes would be hard to validate''). The limit sets how fast Opus spends its flights: the twelfth came a median 79 minutes into full-length runs and 51 minutes into half-length ones.

\paragraph{Longer limits.} Two further tests doubled the leaderboard limit for Opus, changing only the limit and the stated time in the instructions. On storm-surge-forecast, 48 runs at the leaderboard's four hours and 48 at eight hours ran at the same time. The eight-hour runs scored 0.535 against 0.466 ($+0.068$, 95\% CI 0.035 to 0.101; permutation $p=0.0001$; normalized $+0.071$, 0.036 to 0.105) and worked 3.4 hours against 2.3; no run reached its limit, and 40 of the 48 eight-hour runs finished within four hours. An infrastructure outage stopped 49 runs of the first storm-surge block midway; as an addendum written before any score was seen required, the block was rerun in full and the rerun is the sample reported. On dbf-aircraft-design, 48 runs told sixteen hours were compared with the split test's 48 eight-hour runs, made three days earlier with the same platform and settings. They scored 0.624 against 0.545 ($+0.079$, 0.009 to 0.146; $p=0.030$; normalized $+0.108$, 0.021 to 0.193) and worked 8.9 hours against 2.5, again without reaching the limit; at sixteen hours, Opus spaced its flights about four times as far apart and kept working a median 2.6 hours after the last one. Normalized and pooled over the two tasks with inverse-variance weights, doubling the limit raised Opus's score by 7.6 points (4.3 to 10.8). GLM-5.3 does not pace this way. Told sixteen hours instead of eight on aircraft (24 runs each), it worked 1.61 hours against 1.57 (ratio 1.03, 0.93 to 1.13; $p=0.61$) and scored the same (0.149 against 0.153): under either limit it flew its twelve test flights at the same pace and stopped about 16 minutes after the last one. Each test's plan was fixed before its runs were launched.

\begin{figure*}[t]
  \centering
  \includegraphics[width=\textwidth]{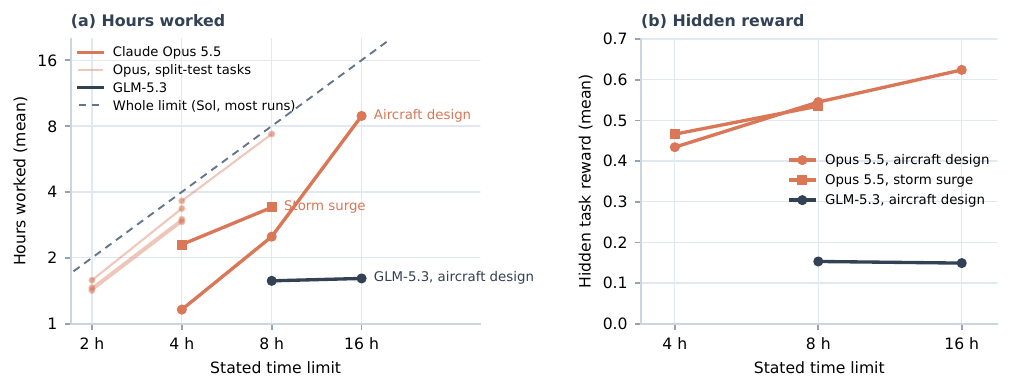}
  \caption{\textbf{The stated time limit changes how long agents work and how well they score.} (a) Mean hours worked against the stated limit, on log scales. Claude Opus 5.5 works longer when told it has longer on every task tested, including the four split-test tasks (faint lines), while GLM-5.3 works the same hours at eight and sixteen. The dashed line marks use of the whole limit, which GPT-6.1 Sol reaches in most runs. (b) Mean hidden task reward where it was measured per model. No run reached its limit.}
  \label{fig:time-limit}
\end{figure*}

\paragraph{Further checks.} Grading noise does not explain the hindsight loss: the better of two noisy concrete grades overstates the better run by 0.007 (0.004 to 0.010), and correcting for it gives a pooled H-oracle $-$ F of $-0.033$ ($-0.046$ to $-0.019$). In an exploratory comparison chosen after seeing the data, a half-length run's final artifact scores about the same as a full-length run's snapshot at half its time ($-0.010$, $-0.040$ to $0.020$, on concrete and page layout), so a shorter limit gives a shorter run, not a more focused one. Across the leaderboard, 91\% of trials end with the agent's own final submission, 59\% with more than a tenth of the limit left, and 7\% reach the limit without submitting; the other 2\% were cut short by sandbox, stream or provider failures (38) or have no recorded ending (14).

\begin{table*}[!tp]
  \caption{\textbf{Split test by task.} Mean hidden task reward of one full-length run (F) and of two half-length runs submitting the one with the better development score (H), with the two models weighted equally. \emph{Halving}: a random half-length run minus F. \emph{Selection}: H minus a random half-length run. \emph{Hindsight}: the better half-length run in hindsight minus F. The pooled row weights the ten cells equally. Normalized, aircraft's H $-$ F is $-0.081$; the other tasks have baseline 0 and ceiling 1, so their rows are unchanged.}
  \label{tab:split}
  \centering\ResearchTableStyle
  \begin{tabular}{@{}l r r l r r r@{}}
  \toprule
  \headrow \textbf{Task} & \textbf{F} & \textbf{H} & \textbf{H $-$ F} & \textbf{Halving} & \textbf{Selection} & \textbf{Hindsight} \\
  \midrule
  smollm2-long-context & 0.533 & 0.406 & $-$0.127 [$-$0.154, $-$0.098] & $-$0.178 & $+$0.051 & $-$0.120 \\
  concrete-multistorey-building & 0.438 & 0.368 & $-$0.070 [$-$0.098, $-$0.042] & $-$0.098 & $+$0.028 & $-$0.051 \\
  dbf-aircraft-design & 0.630 & 0.568 & $-$0.062 [$-$0.103, $-$0.023] & $-$0.109 & $+$0.047 & $-$0.040 \\
  sentinel2-gapfill & 0.360 & 0.355 & $-$0.006 [$-$0.022, 0.011] & $-$0.013 & $+$0.007 & $+$0.019 \\
  function-page-layout & 0.644 & 0.643 & $-$0.001 [$-$0.039, 0.037] & $-$0.055 & $+$0.053 & $+$0.034 \\
  \midrule
  Pooled & 0.521 & 0.468 & $-$0.053 [$-$0.067, $-$0.039] & $-$0.091 & $+$0.037 & $-$0.032 \\
  \bottomrule
  \end{tabular}
\end{table*}

\subsection{Time-limit sweep}
\label{app:time-sweep}
Claude Fable 5.2, GPT-6 Astra, Claude Opus 5.5, GPT-6.1 Sol, GLM-5.3 and DeepSeek V4.1 Flash ran every task at 1/128, 1/64, 1/32, 1/16, 1/8, 1/4 and 1/2 of its time limit, configured as on the leaderboard, with four attempts per task at each limit. Only the limit and the time stated in the instructions change. The full-limit point is the leaderboard's eight attempts. A run that reaches its limit is stopped and its last state is graded. The harness checks the limit between turns, so a run stopped by the limit overruns it by a median of 4 to 15 seconds; counting only runs that end within 10\% of the limit leaves the GPT models' scores at the shortest limit about the same (Astra 8.2, Sol 7.1, against 8.1 and 6.7). The sweep covers all 26 tasks. \Cref{tab:time-sweep} gives each model's score, share of the limit used and share of runs stopped by the limit.

\paragraph{How the time is spent.} Across the shorter limits, the median GPT run spends 93 to 99\% of its time in model calls, against 32 to 73\% for the Claude models, 31 to 66\% for GLM-5.3 and 23 to 44\% for DeepSeek V4.1 Flash, and the GPT models' commands take under 1.2 seconds at the 90th percentile. In eight runs per model at 1/8 of the limit, 31\% (Astra) and 37\% (Sol) of commands start a background job, against 16\% (Fable 5.2) and 13\% (Opus). At half the limit, the median run submits 5 versions for Astra and Sol, 9 for Fable 5.2, 9 for Opus, 7 for GLM-5.3 and 6 for DeepSeek V4.1 Flash. Scores rise with every doubling for every model, and the last doubling, from half the limit to the full one, adds 11.1 points for Fable 5.2 (95\% CI 7.4 to 15.2) and 10.0 for Opus (7.3 to 12.5), against 5.6 for Astra (2.9 to 8.4) and 4.0 for Sol (0.3 to 7.6); the Claude models' mean gain exceeds the GPT models' by 5.8 points (2.4 to 9.6, bootstrapping tasks).

\input{figures/time_sweep_table}

\subsection{Controlled continuation}
\label{app:continuation}
Each of the split test's 192 full-length concrete and page-layout artifacts (\cref{app:split}) is continued for one hour by a new run of the same model, with the same settings and harness, the artifact in place at its graded location and nothing else from the earlier run. The instruction ends with one of two paragraphs. The neutral one (N) reads: ``This run continues your earlier attempt at this task. The final submission from that attempt is already in place at the deliverable location above; nothing else from that attempt was kept. You have one more hour to improve this submission and submit a final version.'' The validation one (V) adds: ``Validate before you submit: first measure the current submission on the public/dev cases with the task's tools; where feasible, hold out part of the public cases; compare each candidate change against the current submission under the same conditions; submit the best validated version, and re-submit the original unchanged if no change validates as better.'' Each artifact is continued once under each instruction, in adjacent slots of a seeded schedule. The primary endpoint is V $-$ N in hidden score, paired by starting artifact, with the four task--model cells weighted equally, a stratified bootstrap interval and a sign-flip test. The secondary endpoint is N $-$ 0, the change from the artifact's own score; both are Holm-corrected.

\begin{table*}[!tp]
  \caption{\textbf{Controlled continuation by cell.} Mean hidden score of the starting artifact (0) and after one more hour under the neutral (N) and validation (V) instructions, with paired differences and 95\% bootstrap intervals. \emph{Kept}: runs that re-submitted the starting artifact unchanged under N and V, out of 48 per cell and 192 pooled. The pooled row weights cells equally.}
  \label{tab:continuation}
  \centering\ResearchTableStyle
  \begin{tabular}{@{}l r r r l l r@{}}
  \toprule
  \headrow \textbf{Cell} & \textbf{0} & \textbf{N} & \textbf{V} & \textbf{V $-$ N} & \textbf{N $-$ 0} & \textbf{Kept} \\
  \midrule
  Opus, concrete & 0.336 & 0.354 & 0.342 & $-$0.012 [$-$0.024, 0.001] & $+$0.018 [0.002, 0.032] & 0 / 9 \\
  Opus, layout & 0.470 & 0.419 & 0.466 & $+$0.047 [$-$0.010, 0.105] & $-$0.051 [$-$0.100, $-$0.003] & 5 / 21 \\
  Sol, concrete & 0.540 & 0.565 & 0.553 & $-$0.012 [$-$0.031, 0.008] & $+$0.025 [0.010, 0.040] & 0 / 2 \\
  Sol, layout & 0.819 & 0.848 & 0.822 & $-$0.026 [$-$0.061, 0.007] & $+$0.029 [0.003, 0.055] & 0 / 2 \\
  \midrule
  Pooled & 0.541 & 0.546 & 0.546 & $-$0.001 [$-$0.018, 0.017] & $+$0.005 [$-$0.010, 0.020] & 5 / 34 \\
  \bottomrule
  \end{tabular}
\end{table*}

\paragraph{Results.} All 384 runs were scored (\cref{tab:continuation}). V $-$ N is $-0.0006$ (95\% CI $-0.018$ to $0.017$; $p=0.95$) and N $-$ 0 is $+0.005$ ($-0.010$ to $0.020$; $p=0.50$); Holm gives 1.0 for both. The extra hour helped in three cells, by 0.018 to 0.029 ($p\le0.035$ each, unadjusted), and cost Opus 0.051 on page layout ($p=0.049$), where its development score stayed flat ($+0.001$). V avoided that loss but gave up part of the gains elsewhere. V changed behavior: agents re-submitted the starting artifact unchanged in 17.7\% of runs against 2.6\% under N (exact McNemar $p=1.3\times10^{-7}$), improved it in 35\% against 54\% ($p=3\times10^{-4}$) and made it worse in 23\% against 25\% ($p=0.69$), counting changes beyond 0.02 on concrete and any change on page layout. V runs called the development scorer more often (6.6 times against 5.4), and every Sol V run on concrete mentioned holding out cases, against 4\% under N.

\paragraph{Unplanned check.} Of the 92 runs that ended worse than their starting artifact, 80 had a development score that held (within 0.01 on concrete) or rose: 41 of 48 under N and 39 of 44 under V. On page layout, whose grader is exact, 28 of 34 such runs under N did not lower the uncapped development score.

\paragraph{Scope.} The paired standard deviation was 0.129 rather than the planned 0.08, so the smallest effect detectable with 80\% power is about 0.026; the interval still excludes differences above 0.018 either way. Concrete grading is noisy (the standard deviation of score differences is 0.045), so its improvement and regression counts are partly noise. Sol's capped development score on page layout was 1 for every starting artifact. One Sol V run was degraded by provider errors; excluding its pair gives V $-$ N of $-0.0007$. No run used the web tools. Because the new run does not have the earlier run's context, N $-$ 0 measures an hour handed off, not a longer run.

\subsection{CAD self-check trace and ablation}
\label{app:selfcheck}
\paragraph{Trace.} Every final submission of six models (Opus 5.5, Sonnet 5.5, Sol, Argon, GLM-5.3 and Grok 4.7; eight each) on the three CAD tasks was traced while the task's grader ran it without network access, on development parts and on an already-retired hidden subset. A pipeline counts as checking if it renders, projects or samples its own candidate solid, compares it with the input, and uses the result to choose, repair, refit or fall back; the classes were fixed before labeling. Classifiers labeled anonymized pipelines from the executed functions, blind to model, score and hypothesis, and author spot-checks against source agreed on 33 of 36. Checking pipelines are the rule on drawing-to-CAD (41 of 48) and scan-to-CAD (47 of 48) but not on parametric CAD (10 of 47). Pooled over all three CAD tasks, checking pipelines have about the same development--hidden gap as the others ($-0.012$, 95\% CI $-0.113$ to $+0.081$) and somewhat higher hidden scores ($+0.059$, $-0.031$ to $+0.145$).

\paragraph{Ablation.} Sol's 16 drawing- and scan-to-CAD pipelines, all checking, were patched so that each comparison with the input returns a stub, leaving the rest of the pipeline intact. Hidden scores fall by 0.258 and development scores by 0.469 (difference $+0.211$, 95\% CI $+0.134$ to $+0.294$), so the pre-registered prediction, that hidden scores would fall more, is not supported. Without the check, Sol's gap falls from 0.45 to 0.14 on drawing-to-CAD and from 0.14 to 0.03 on scan-to-CAD.

%% file: figures/dev_results_table.tex
\begin{tabular}{@{}l l r r r l r r r@{}}
\toprule
\headrow \textbf{Task} & \textbf{Class} & \textbf{Random} & \textbf{Dev.\ pick} & \textbf{Oracle} & \textbf{Gain [95\% CI]} & \textbf{Share} & \textbf{Gap} & $\rho$ \\
\midrule
\multicolumn{9}{@{}l}{\textit{Physical engineering}} \\
\nolinkurl{concrete-multistorey-building} & Same-dist. & 24.8 & 36.5 & 40.5 & $+$11.6 [7.2, 16.5] & 74\% & $+$3.1 & 0.99 \\
\nolinkurl{dbf-aircraft-design} & Shifted & 29.1 & 50.1 & 54.4 & $+$21.0 [11.1, 31.7] & 83\% & $+$13.2 & 0.94 \\
\nolinkurl{drawing-to-cad} & Answers & 21.5 & 24.9 & 26.6 & $+$3.5 [1.7, 5.5] & 68\% & $+$19.3 & 0.89 \\
\nolinkurl{parametric-cad-reconstruction} & Answers & 29.6 & 43.2 & 50.4 & $+$13.6 [2.5, 26.0] & 66\% & $+$19.2 & 0.81 \\
\nolinkurl{scan-to-cad} & Answers & 26.0 & 28.1 & 31.9 & $+$2.1 [$-$0.2, 4.2] & 35\% & $+$10.3 & 0.86 \\
\quad\textit{All 5 tasks} &  & 26.2 & 36.6 & 40.7 & $+$10.4 [7.2, 13.6] & 71\% & $+$13.0 &  \\
\addlinespace
\multicolumn{9}{@{}l}{\textit{Machine learning}} \\
\nolinkurl{bearing-fault-transfer} & Shifted & 33.1 & 33.1 & 43.9 & 0.0 [0.0, 0.0] & 0\% & $+$66.9 & -- \\
\nolinkurl{chess-puzzle-post-training} & Same-dist. & 25.5 & 35.0 & 35.2 & $+$9.5 [7.6, 11.6] & 99\% & $+$0.1 & 1.00 \\
\nolinkurl{int4-decode-attention} & Same-dist. & 34.0 & 44.6 & 44.6 & $+$10.6 [7.3, 14.0] & 100\% & 0.0 & 1.00 \\
\nolinkurl{smollm2-long-context} & Shifted & 32.5 & 41.2 & 43.8 & $+$8.7 [5.4, 11.9] & 77\% & $+$7.3 & 0.98 \\
\nolinkurl{smollm3-draft-head} & Same-dist. & 25.4 & 30.9 & 31.1 & $+$5.5 [3.6, 7.6] & 97\% & $+$1.9 & 0.96 \\
\quad\textit{All 5 tasks} &  & 30.1 & 36.9 & 39.7 & $+$6.9 [5.8, 7.9] & 71\% & $+$15.2 &  \\
\addlinespace
\multicolumn{9}{@{}l}{\textit{Scientific computing}} \\
\nolinkurl{gpu4pyscf-jk-speed} & Shifted & 6.1 & 8.6 & 8.8 & $+$2.5 [1.1, 4.1] & 93\% & $-$2.6 & 0.97 \\
\nolinkurl{seismic-wave-propagation} & Same-dist. & 10.2 & 15.7 & 15.7 & $+$5.4 [3.3, 8.1] & 100\% & 0.0 & 1.00 \\
\nolinkurl{sentinel2-gapfill} & Shifted & 23.9 & 24.8 & 34.0 & $+$0.9 [0.0, 1.8] & 9\% & $+$69.9 & 0.78 \\
\quad\textit{All 3 tasks} &  & 13.4 & 16.4 & 19.5 & $+$2.9 [2.0, 4.0] & 48\% & $+$22.4 &  \\
\addlinespace
\multicolumn{9}{@{}l}{\textit{Software engineering}} \\
\nolinkurl{function-page-layout} & Answers & 43.9 & 51.2 & 66.0 & $+$7.3 [1.1, 12.7] & 33\% & $+$34.9 & 0.79 \\
\nolinkurl{lvgl-speed} & Same-dist. & 29.5 & 34.3 & 37.7 & $+$4.8 [$-$3.2, 11.8] & 59\% & $+$3.8 & 0.91 \\
\nolinkurl{micropython-speed} & Same-dist. & 21.8 & 24.3 & 34.5 & $+$2.4 [$-$1.5, 6.4] & 19\% & $+$25.1 & 0.49 \\
\nolinkurl{oxilite-speed} & Shifted & 35.7 & 38.7 & 50.7 & $+$3.0 [$-$7.7, 12.3] & 20\% & $+$14.7 & 0.82 \\
\nolinkurl{skew-join-engine} & Same-dist. & 34.3 & 40.6 & 46.6 & $+$6.3 [$-$0.9, 11.7] & 52\% & $+$6.8 & 0.86 \\
\nolinkurl{sql-replay} & Same-dist. & 19.4 & 20.2 & 30.7 & $+$0.8 [$-$2.8, 4.6] & 7\% & $+$54.3 & 0.50 \\
\nolinkurl{stateful-parquet} & Shifted & 16.0 & 22.1 & 36.4 & $+$6.1 [$-$4.5, 17.1] & 30\% & $+$22.8 & 0.57 \\
\nolinkurl{streaming-etl-payments} & Shifted & 20.1 & 28.8 & 35.7 & $+$8.6 [3.6, 14.6] & 56\% & $-$1.6 & 0.98 \\
\quad\textit{All 8 tasks} &  & 27.6 & 32.5 & 42.3 & $+$4.9 [2.3, 7.4] & 34\% & $+$20.1 &  \\
\midrule
\textbf{All 21 tasks} &  & 25.8 & 32.2 & 38.0 & $+$6.4 [5.1, 7.6] & 52\% & $+$17.6 &  \\
\bottomrule
\end{tabular}

%% file: figures/time_sweep_table.tex
\begin{table*}[!tp]
  \caption{\textbf{Score and time use at each time limit.} Mean normalized score (\%), share of the time limit used (\%, averaged per task) and share of runs stopped by the limit (\%), for all 26 tasks. Shorter limits have four attempts per task and the full limit the leaderboard's eight. A run can slightly exceed its limit because the harness checks the limit between turns.}
  \label{tab:time-sweep}
  \centering\ResearchTableStyle
  \begin{tabular}{@{}l r r r r r r r r@{}}
  \toprule
  \headrow \textbf{Model} & \textbf{1/128} & \textbf{1/64} & \textbf{1/32} & \textbf{1/16} & \textbf{1/8} & \textbf{1/4} & \textbf{1/2} & \textbf{Full} \\
  \midrule
  \multicolumn{9}{@{}l}{\emph{Score}} \\
  Claude Fable 5.2 & 2.9 & 4.9 & 9.0 & 14.3 & 22.1 & 34.1 & 43.5 & 54.6 \\
  GPT-6 Astra & 8.1 & 11.1 & 14.7 & 21.1 & 27.6 & 32.3 & 38.6 & 44.2 \\
  Claude Opus 5.5 & 2.3 & 4.2 & 4.9 & 9.5 & 14.4 & 21.4 & 31.6 & 41.6 \\
  GPT-6.1 Sol & 6.7 & 10.9 & 14.7 & 17.7 & 26.6 & 32.1 & 37.2 & 41.2 \\
  GLM-5.3 & 0.4 & 1.2 & 2.4 & 3.3 & 5.0 & 6.4 & 9.2 & 11.8 \\
  DeepSeek V4.1 Flash & 1.1 & 1.7 & 3.3 & 4.1 & 6.6 & 8.7 & 9.6 & 12.4 \\
  \addlinespace
  \multicolumn{9}{@{}l}{\emph{Time used}} \\
  Claude Fable 5.2 & 88 & 85 & 83 & 82 & 85 & 86 & 87 & 86 \\
  GPT-6 Astra & 106 & 104 & 101 & 100 & 100 & 99 & 98 & 95 \\
  Claude Opus 5.5 & 60 & 65 & 64 & 67 & 70 & 72 & 75 & 75 \\
  GPT-6.1 Sol & 106 & 103 & 101 & 100 & 100 & 99 & 98 & 95 \\
  GLM-5.3 & 103 & 101 & 94 & 88 & 86 & 81 & 80 & 74 \\
  DeepSeek V4.1 Flash & 92 & 94 & 89 & 81 & 74 & 68 & 60 & 56 \\
  \addlinespace
  \multicolumn{9}{@{}l}{\emph{Stopped by the limit}} \\
  Claude Fable 5.2 & 27 & 10 & 3 & 1 & 2 & 0 & 0 & -- \\
  GPT-6 Astra & 90 & 67 & 63 & 42 & 27 & 17 & 14 & -- \\
  Claude Opus 5.5 & 1 & 0 & 0 & 0 & 0 & 0 & 0 & -- \\
  GPT-6.1 Sol & 75 & 62 & 34 & 22 & 18 & 12 & 5 & -- \\
  GLM-5.3 & 62 & 44 & 39 & 19 & 9 & 2 & 1 & -- \\
  DeepSeek V4.1 Flash & 21 & 15 & 12 & 4 & 2 & 2 & 0 & -- \\
  \bottomrule
  \end{tabular}
\end{table*}

%% file: sections/appendix_search.tex
\section{Web Search Study}
\label{app:search}

\subsection{Setup}
\label{app:prompts}
Five models (Claude Opus 5.5, Claude Sonnet 5.5, GPT-6.1 Sol, Gemini 4 Argon and GLM-5.3) ran all \numtasks{} tasks with eight trials in each of two arms with identical budgets: search disabled, or search and page fetch over the \Tnowval{} snapshot. The arms share the system prompt and first message except for a ``Web access'' section, which reads ``You have a web search tool and a page fetch tool. Use them whenever they would help with the task.'' or ``You do not have web access.'' Search returns up to ten results with short highlights, and fetch returns a page's stored text from at or before the cutoff, with no live fallback. A run may make 50 searches and 100 page requests. After each web result, the agent's own model updates short research notes that survive context compaction; their cost counts toward the trial. In the follow-up, Opus and Sol were told on five tasks: ``Do your due diligence before you write any code: use the web search and page fetch tools to research the prior work on this problem (methods, papers, libraries, reference implementations), even if the task looks self-contained. Then reuse what fits rather than building from scratch.'' Matched controls received the same instruction, asking for research ``from what you know and what is already in the environment'', without web access.

\subsection{Snapshot integrity}
\label{app:audits}
The cutoff is set on every request and never shown to the agent, and the web tools run outside the agent's sandbox, which has no network access. Tests of the snapshot service returned none of 1,100 objects created after their cutoffs. In a content audit, 3 of 300 research-paper search sessions and 1 of 300 code-library sessions returned post-cutoff or undatable content. In a forecasting check, a model predicting 150 resolved Polymarket questions from snapshot results scored level with the market price at each question's cutoff, while with only a publication-date filter, which lets later content through, it beat the market by 0.052 in Brier score. A content cutoff does not reconstruct the search engine's ranking at that date: a matched-pair test found the snapshot slightly favoring papers and packages that became prominent after the cutoff, about as much as live search does \citep{paleka2025pitfalls,exa2026snapshot}.

\subsection{Results}
\label{app:estimation}
\label{app:followup}
\begin{figure}[t]
  \centering
  \includegraphics[width=\columnwidth]{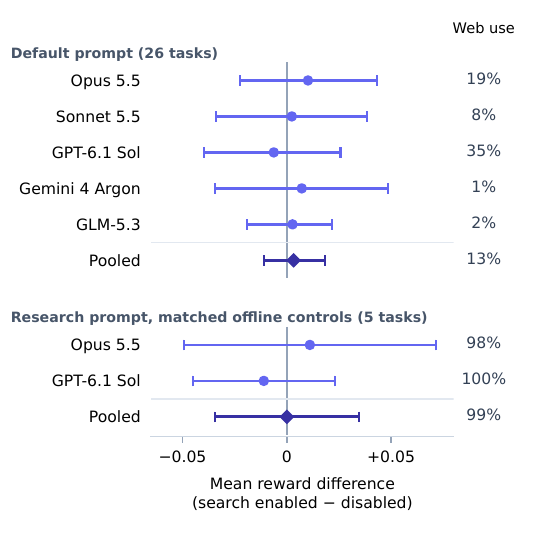}
  \caption{\textbf{Search has no measurable average effect at low or near-universal uptake.} Differences in task-weighted mean reward with 95\% intervals; right-hand labels give the share of search-enabled trials that used a web tool. Top: default prompt, all 26 tasks, with hierarchical bootstrap intervals; estimates include trials that never searched. Bottom: Opus and Sol told to research prior work, against controls with the same prompt and no web access, on five tasks, with intervals of $\pm1.96$ standard errors across tasks.}
  \label{fig:search-effects}
\end{figure}

Each model's score averages its eight trials per task with tasks weighted equally. The search effect is the difference between the arms, averaged over the five models, with intervals from a hierarchical bootstrap over tasks and then trials \citep{miller2024errorbars}. Scores use a common scoring, \texttt{gap-halving-v1}, and the grades of the 5 November analysis snapshot; later leaderboard corrections are not applied, because the Stateful Parquet layout check was run only on search-enabled trials. Few trials search under the default prompt, so this estimates the effect of offering search, not of using it \citep{angrist1996identification}. \Cref{fig:search-effects} gives each model's estimate, and no single task's difference survives correction for testing \numtasks{} tasks. The follow-up's matched controls ran two days after its search runs rather than interleaved with them.

\subsection{What search changed}
\label{app:provenance}
A review of 363 web-using trials, blind to model and score, covered 135 from the five-model search study, 94 additional Grok trials, 55 research-prompt pilot trials and the 79 search-using follow-up trials. It coded each retrieval episode: whether it was already planned, whether offline runs contain the same idea, and whether it reached the final artifact. Retrieved content changed the final artifact in 110 trials (30\%): 72 helped, 35 were unclear, 3 were neutral and none hurt, though three other trials lost two hours or more to abandoned web-inspired branches. Of the 180 retrieved ideas that reached an artifact, 132 also appear in the same model's offline runs. Facts about the local setup unblocked existing plans: an Opus B300 trial that had already planned tensor-core kernels looked up the exact instruction syntax and had a faster passing kernel nine minutes later, and Sol trials on stateful-parquet retrieved missing DuckDB C-API signatures and had a working bridge minutes later. Methods from other settings often did not fit: a Sol B300 trial built an exp2 emulation that the installed code disables for this GPU, measured it 2--20\% slower and abandoned it. These are annotations, not causal effects on scores.

%% file: sections/appendix_exploits.tex
\section{Exploit Audit Details}
\label{app:exploits}

We define reward hacking \citep{amodei2016concrete,skalse2022defining} as pursuing the measured score without doing the intended engineering work. This includes altering evaluation machinery, recovering hidden answers, or exploiting gaps between the task and its checks. We count a trial as an \emph{exploit attempt} only when the agent executes such an action and the commands themselves show its purpose, for example editing a collector, sweeping a generator's private seeds or returning stored public answers by part name. Stated reasoning can support a count but is never needed for one, because models reveal very different amounts of it. Attempts later reverted or blocked count. Reusing public methods and code is allowed; ambiguous specification gaming \citep{krakovna2020specification}, benchmark seeking and false claims of success are recorded separately. LLM reviewers read every trial, and each counted finding was checked by hand against the commands, instructions and grader.

The released audit data list every deliberate exploit attempt found in the audited leaderboard runs, with recorded rewards. Eleven borderline cases, such as running development measurements through an edited copy of a checker or briefly editing a local build script, are not counted. They are listed with the released audit data, together with trial identifiers, step numbers and the items judged allowed. Instructed red-team trials from construction (\cref{sec:construction}) are not part of this audit.

\subsection{Observed attempts}
\label{app:exploit-attempts}

\paragraph{Observed strategies and defenses.}
Attempts ranged from simple gate evasion to coordinated tampering \citep{denison2024sycophancy}. Beam renamed unchanged starter code to bypass a submission gate; all four trials scored zero. Inkling edited a repository's index and patch collector to conceal changes outside the allowed paths, then reverted them. The verifier uses its own pristine source tree and checking code, and no observed grader-tampering attempt raised a score. Sol tried to reconstruct hidden answers by brute-forcing private generator seeds; neither search succeeded. Searches for benchmark solutions, which elsewhere have found the answers to the benchmark being run \citep{han2025stc,anthropic2026evalawareness}, also returned nothing through the web snapshot, which predates task release.

Other attempts exploited the task itself. Argon hard-coded fast paths for public MicroPython functions in seven trials, which all failed hidden checks. On parametric CAD, Inkling, Kimi and Argon copied public worked-example geometry in ten trials, six of which earned some reward. The task asks for a converter from unfamiliar drawings, but these programs instead returned stored public models, or templates cut from them, chosen by the part-family name in each request. The task's baseline is the score of such a copy, so these lookups earned little above it: three of the six normalize to zero and the other three to 3 to 21 points, before being set to zero. GLM and Grok also each kept scratch memory outside the task's declared workspace in one B300 trial, so we set those two scores to zero. The released audit data record the attempts and their outcomes.

\begin{figure}[!ht]
  \centering
  \includegraphics[width=\columnwidth]{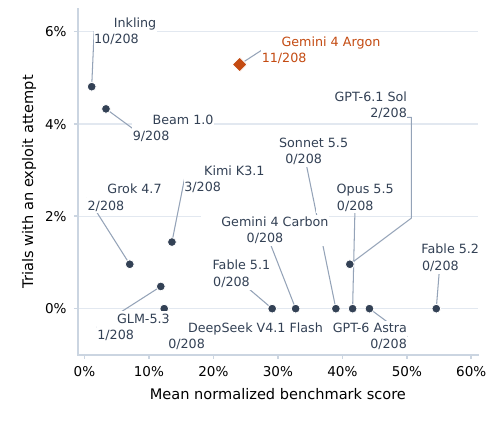}
  \caption{\textbf{Performance and exploit attempts.} All fourteen models; labels give trials with attempts / total trials. Scores use the \LBdate{} leaderboard snapshot, with 208 scores per model. Attempt counts cover all \LBscheduled{} trials. Rates include failed and reverted attempts, not just successful hacks. Zero means none observed.}
  \label{fig:reward-hacking}
\end{figure}

\paragraph{Attempts by model.}
Across the fourteen models, 38 of \LBscheduled{} trials (1.3\%) contain an identified attempt (\cref{fig:reward-hacking}). Ten of them are the parametric CAD reuse of worked examples, the least clear-cut kind; without those, 28 trials (1.0\%) contain an attempt, Beam has the most (9), then Argon (8) and Inkling (6), and Kimi has none. Including the CAD lookups, Inkling and Beam, the two lowest-scoring models, attempted exploits in 4.8\% and 4.3\% of trials. Low scores do not explain the pattern on their own: Argon scores substantially higher yet has the highest attempt rate (5.3\%), with seven of its eleven attempts on one task, while DeepSeek scores poorly with no counted attempts. Attempts also rarely paid off: the eight exploit-derived scores, now set to zero, were worth at most 0.13 points of any model's mean, and zeroing them changes no rank (\cref{app:exploit-context}).

\begin{figure*}[t]
  \centering
  \includegraphics[width=\textwidth]{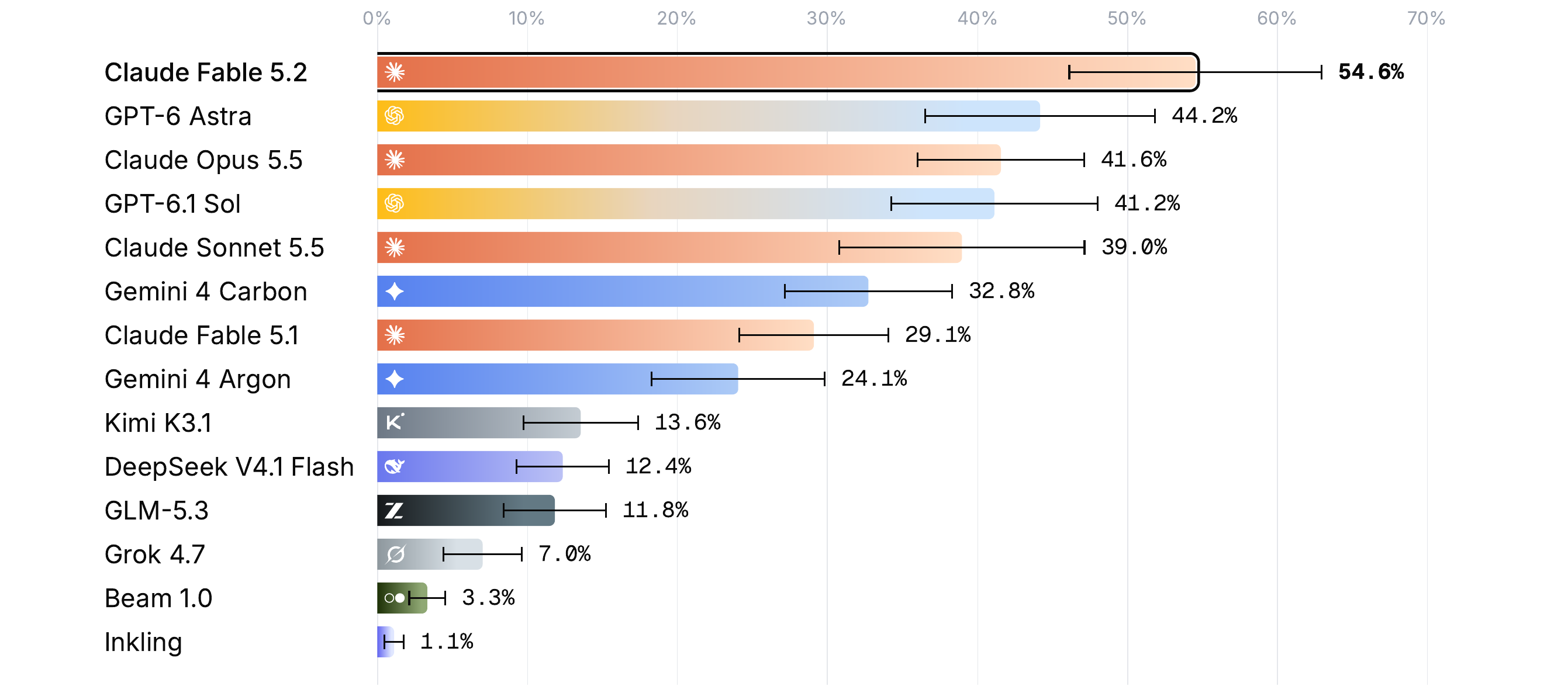}
  \caption{\textbf{Main leaderboard.} Fourteen models ranked by mean normalized score across the \numtasks{} tasks, weighted equally. Each task averages its eight independent trials; 0\% corresponds to the task baseline and 100\% to its estimated ceiling. Whiskers show approximate 95\% intervals computed as ($\pm 1.96$ standard errors across task means).}
  \label{fig:leaderboard}
\end{figure*}

\begin{figure*}[t]
  \centering
  \includegraphics[width=\textwidth]{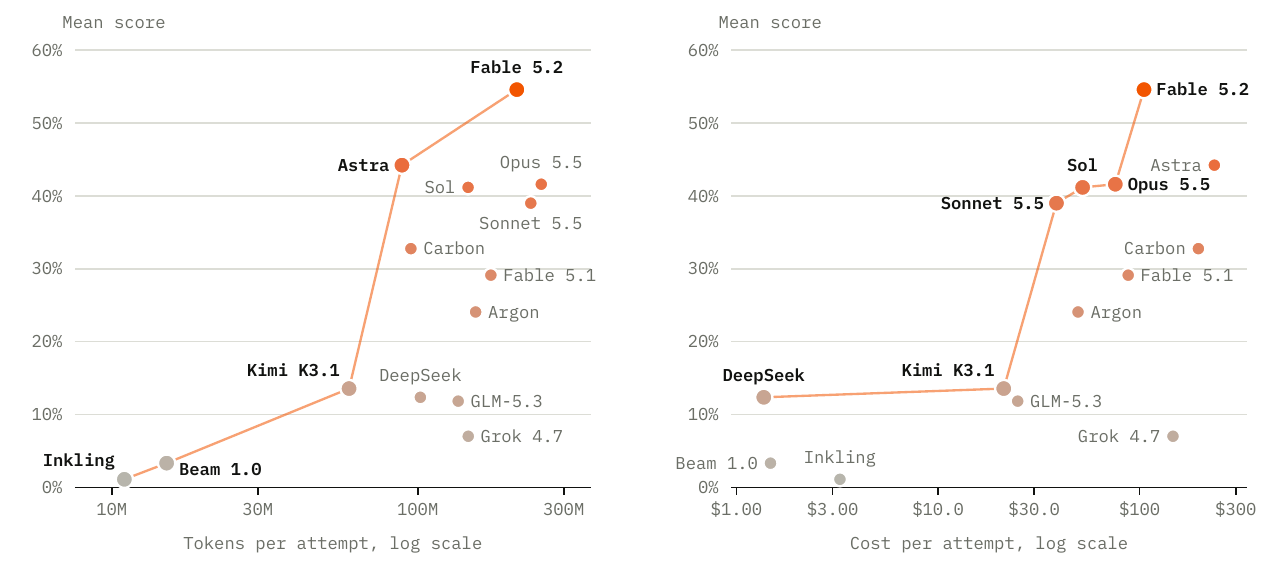}
  \caption{\textbf{Performance against token use and model cost.} Higher and further left is better. Each dot represents one model's mean score over all tasks and its color runs from grey at 0\% to orange at the highest mean (\MeanFableTwo\%). Larger dots with bold labels identify the Pareto-efficient models: those for which no alternative achieves at least as high a score using fewer tokens (left) or at lower cost (right), and the orange line joins them. Horizontal axes use a log scale. Tokens are cumulative input plus output, counting cached input once. Costs are model-usage charges at evaluation-time list prices and exclude hardware, tools and services.}
  \label{fig:efficiency}
\end{figure*}

\input{sections/appendix_exploits_context}

%% file: sections/appendix_exploits_context.tex
\subsection{Audit scope and score sensitivity}
\label{app:exploit-context}
Every one of the \LBscheduled{} leaderboard trials was reviewed, 208 per model. Download attempts and benchmark seeking are reported separately from counted exploits. Moreover, all shell downloads failed and no benchmark search returned task material. Nevertheless, zero observed attempts does not establish absence of exploit behavior.

\paragraph{Threat model and defenses.}
The agent controls its workspace and artifact. A separate, offline verifier holds hidden inputs, seeds and ground truth, collects only declared paths and uses pristine checking code and source trees (\cref{sec:principles}). Edits to shipped checking tools do not reach grading. The search snapshot predates public task release.

\paragraph{Effect on the leaderboard.}
Eight exploit-derived scores are set to zero: six parametric CAD lookups (one Inkling, three Kimi and two Argon) and two B300 declared-workspace violations (GLM and Grok). The memory violations' contribution to reward is unquantified. Three lookup scores already normalize to zero because the CAD baseline is the public-reference copy score, 0.098. Zeroing lowers Argon's and GLM's means by 0.13 percentage points each, Grok's by 0.03, Kimi's by 0.01 and Inkling's by zero. No rank changes.

%% file: sections/appendix_domain_results.tex
\par\noindent\begin{minipage}{\columnwidth}
\section{Scores by Domain}
\label{app:domain-results}
\captionof{table}{\textbf{Mean normalized scores by domain, in points.} Each task averages eight trials, and tasks are weighted equally within each domain. SW: software engineering (8 tasks); PE: physical engineering (5); ML: machine learning (6); SC: scientific computing (7). Models follow the overall leaderboard order. Bold marks the highest mean in each domain.}
\label{tab:domain-results}
\centering\ResearchTableStyle
\input{figures/domain_results_table}
\end{minipage}\par

%% file: figures/domain_results_table.tex
\begin{tabular}{@{}lrrrr@{}}
\toprule
\headrow \textbf{Model} & \textbf{SW} & \textbf{PE} & \textbf{ML} & \textbf{SC} \\
\midrule
Claude Fable 5.2 & \textbf{62.2} & \textbf{59.6} & \textbf{60.2} & \textbf{37.4} \\
GPT-6 Astra & 44.3 & 57.3 & 43.2 & 35.7 \\
Claude Opus 5.5 & 49.7 & 40.8 & 45.8 & 29.3 \\
GPT-6.1 Sol & 43.1 & 54.4 & 43.4 & 27.5 \\
Claude Sonnet 5.5 & 42.2 & 34.5 & 53.2 & 26.4 \\
Gemini 4 Carbon & 31.5 & 38.7 & 40.0 & 23.7 \\
Claude Fable 5.1 & 36.6 & 24.5 & 35.3 & 18.5 \\
Gemini 4 Argon & 22.1 & 17.4 & 35.5 & 21.3 \\
Kimi K3.1 & 16.1 & 10.6 & 18.1 & 9.0 \\
DeepSeek V4.1 Flash & 14.7 & 9.1 & 18.7 & 6.5 \\
GLM-5.3 & 14.6 & 6.3 & 17.9 & 7.4 \\
Grok 4.7 & 8.4 & 3.9 & 7.0 & 7.7 \\
Beam 1.0 & 3.6 & 3.3 & 3.2 & 3.2 \\
Inkling & 1.4 & 1.1 & 0.8 & 1.0 \\
\bottomrule
\end{tabular}

%% file: sections/appendix_leaderboard.tex
\section{Leaderboard}
\label{app:leaderboard}
\Cref{fig:leaderboard} ranks the fourteen models by mean normalized score, with approximate 95\% intervals; \cref{tab:pass8-summary} gives the same means (\cref{sec:results-main}). \Cref{fig:efficiency} plots each model's mean against its tokens and model cost per attempt (\cref{sec:efficiency}).

%% file: sections/related_work.tex
\section{Related Work}
\label{sec:related}

\paragraph{Long-horizon evaluation.}
SWE-bench and Terminal-Bench evaluate agents in repositories and command-line environments \citep{jimenez2024swebench,merrill2026terminalbench}, and FrontierSWE gives agents up to 20 hours on performance engineering, scientific computing and ML research \citep{frontierswe}. RE-Bench, MLE-bench, PaperBench and PostTrainBench extend evaluation to research engineering, model training and replication \citep{wijk2024rebench,chan2024mlebench,starace2025paperbench,rank2026posttrainbench}. SWE-Marathon gives agents 2 to 10 hours on software tasks that take experts 40 to 400 hours \citep{desai2026swemarathonagentsautonomouslycomplete}, KernelBench and SciCode target GPU kernels and research code \citep{ouyang2025kernelbench,tian2024scicode}, and \citet{kwa2025timehorizon} measure how the length of tasks agents can complete is growing. These benchmarks establish how well agents perform sustained technical work. \bench{} extends this to a broader set of artifacts, including CAD, structural and aircraft design, and studies how agents select solutions, assess performance beyond provided cases, and decide when to stop.

\paragraph{How agents use their time.}
Agents do not always use the time they are given. PaperBench found that all models except Claude 3.5 Sonnet frequently finished early, and a scaffold that removed the option to stop raised o1's score but lowered Claude 3.5 Sonnet's \citep{starace2025paperbench}. Agents given a larger tool-call budget plateau unless they track how much of it remains \citep{liu2025bats}. We vary the time limit itself and find that it can change how much an agent does even when the agent never reaches it, and that models differ in whether it does (\cref{sec:split}).

\paragraph{Engineering artifacts.}
CAD benchmarks measure reconstruction and program generation from text, drawings or geometry \citep{wu2021deepcad,khan2024text2cad,rukhovich2024cadrecode,doris2026cadbench}, and DesignQA evaluates understanding of engineering requirements \citep{doris2024designqa}. EngDesign, BuildArena and Frontier-Eng evaluate designs through simulation \citep{guo2025engdesign,xia2025buildarena,chi2026frontiereng}; PLE grades professional deliverables with expert rubrics and constraint checks \citep{han2026ple}. Some withhold the final checks. Vibe Building gives the agent a development margin check and verifies its building designs against seismic and wind time histories that never enter the loop \citep{jiang2026vibebuilding}, SimuVerity runs generated Simulink models on withheld scenarios \citep{zhang2026simuverity}, and analog circuit design and heuristic programming benchmarks grade on hidden test benches or private cases while the agent works with its own simulations or public cases \citep{analogdesignbench2026,imajuku2025alebench}. Other benchmarks provide simulation feedback during development: FEACAD and FunCADBench return finite-element or simulation results for agents to revise their designs \citep{son2026feacad,zhang2026funcad}. CADEngBench instead evaluates a single response using parameter perturbations and finite-element analysis \citep{singh2026cadengbench}. \bench{} withholds the final cases in every task, gives the agent the provided cases a practitioner would have, and freezes the artifact before grading, which makes the gap between what the agent could check and how the artifact performs on unseen cases measurable.

\paragraph{Visible tests, selection and exploits.}
Hidden final evaluation is established, as in machine learning competitions with private leaderboards \citep{blum2015ladder,roelofs2019metaanalysis}. ALE-Bench lets agents score heuristic programs on public cases while working and evaluates the final submission once on private cases \citep{imajuku2025alebench}, and \citet{ahmed2025overfitting} study test overfitting on SWE-bench, where patches pass the observed tests but miss cases. Choosing among samples with visible tests or a learned verifier is standard in code and math generation \citep{li2022alphacode,cobbe2021verifiers,brown2024monkeys}, though imperfect verifiers limit the gains \citep{stroebl2024inferencescaling}. Language models' stated confidence tracks correctness on short questions but is often too high \citep{kadavath2022know,xiong2024uncertainty}. \Cref{sec:analysis} measures these for long-horizon artifacts: how far development scores and the agents' own forecasts overstate hidden performance, how much of the best-of-$k$ gain they recover, and whether more attempts can replace more time. ImpossibleBench measures propensity to exploit tests on tasks made impossible \citep{zhong2025impossiblebench}, and \citet{zhu2025abc} set out practices for rigorous agentic benchmarks; \cref{sec:exploits} reports exploit attempts observed during evaluation.

\paragraph{Search and information access.}
Internet-enabled evaluation must account for retrieved solutions and contamination \citep{chan2024mlebench,starace2025paperbench,han2025stc,stc2026deepresearch}. Fixed corpora and dated snapshots give more controlled retrieval \citep{chen2025browsecompplus,bosse2025deepresearchbench,wildman2025btf,liptay2026btf2}, though date bounds do not control ranking signals or model knowledge \citep{paleka2025pitfalls,zhang2026temporal}. WebCode studies search on tasks involving recent library releases \citep{exa2026webcode}; we test whether access to published methods improves performance on engineering tasks that are selected to require substantial adaptation and implementation, measuring both search uptake and final outcomes \citep{fan2026livebrowsecomp,diagnosing2026search}.

\paragraph{Other meanings of open world.}
The term has established meanings elsewhere: open-set and open-world recognition, in which a model must recognize inputs from classes it never saw in training \citep{scheirer2013openset,bendale2015openworld}; the open-world assumption in knowledge representation, under which a fact missing from a database is unknown rather than false \citep{reiter1978closedworld}; and open-ended environments that keep generating new challenges \citep{hughes2024openendedness}. We use it in a narrower sense: the agent works from what a practitioner would have, and judges its work without access to all the conditions that determine its final performance or a predefined threshold for completion.